\documentclass{article} 
\usepackage{iclr2027_conference,times}

\usepackage{hyperref}
\usepackage{url}
\usepackage{algorithm}
\usepackage{algpseudocode}
\usepackage{multirow}
\usepackage{xcolor}
\usepackage{graphicx}
\usepackage{wrapfig}
\usepackage{amsmath}
\usepackage[most]{tcolorbox}
\usepackage{xcolor}
\usepackage{tabularx}
\usepackage{amssymb}
\usepackage{amsmath,amssymb}
\usepackage{algorithm}
\usepackage{algpseudocode}
\usepackage{listings}
\usepackage{tcolorbox}
\tcbuselibrary{breakable}

\makeatletter
\AtBeginDocument{\@ifpackageloaded{lineno}{\nolinenumbers}{}}
\makeatother

\usepackage{booktabs,multirow,graphicx}
\usepackage{xcolor}
\usepackage{colortbl}

\definecolor{caterResultRow}{HTML}{EDF4FC}
\definecolor{caterResultGain}{HTML}{67AE6E}
\definecolor{caterResultDrop}{HTML}{B34A44}
\providecommand{\caterGain}[1]{\textcolor{caterResultGain}{#1}}
\providecommand{\caterDrop}[1]{\textcolor{caterResultDrop}{#1}}
\providecommand{\retacCell}[1]{\cellcolor{caterResultRow}#1}

\providecommand{\retacUp}[1]{}
\providecommand{\retacDown}[1]{}
\providecommand{\retacFlat}{}
\renewcommand{\retacUp}[1]{{\tiny\caterGain{(+#1)}}}
\renewcommand{\retacDown}[1]{{\tiny\caterDrop{(-#1)}}}
\renewcommand{\retacFlat}{{\tiny(0.00)}}

\usepackage{array,multirow,graphicx}
\newlength{\retacBaseWidth}
\newlength{\retacMethodWidth}
\newlength{\retacMetricWidth}
\newcommand{\retacTableSetup}{%
  \scriptsize
  \setlength{\tabcolsep}{0.8pt}%
  \renewcommand{\arraystretch}{1.05}%
  \renewcommand{\retacUp}[1]{\,{\caterGain{(+##1)}}}%
  \renewcommand{\retacDown}[1]{\,{\caterDrop{(-##1)}}}%
  \renewcommand{\retacFlat}{\,(0.00)}%
  \settowidth{\retacBaseWidth}{GPT-5.5}%
  \settowidth{\retacMethodWidth}{Gen. Plan.}%
  \settowidth{\retacMetricWidth}{\textbf{00.00\retacUp{00.00}}}%
}

\definecolor{caterAlgVeri}{HTML}{386A94}
\definecolor{caterAlgPlan}{HTML}{76528E}
\definecolor{caterAlg}{HTML}{AA542D}
\definecolor{caterAlgReason}{HTML}{266210}

\definecolor{darkblue}{RGB}{84, 112, 198}
\definecolor{lightgreen}{RGB}{145, 204, 117}
\definecolor{lightyellow}{RGB}{250, 200, 88}
\definecolor{lightred}{RGB}{238, 102, 102}
\definecolor{lightblue}{RGB}{115, 192, 222}
\definecolor{phasecolor}{RGB}{196,103,74}
\newtcolorbox{promptbox}[2][Prompt]{
colback=black!5!white,
arc=5pt, 
boxrule=0.5pt,
fonttitle=\bfseries,
title=#1, 
before upper={\small}, fontupper=\fontfamily{ptm}\selectfont,
colframe=#2,
}

\tcbuselibrary{skins,breakable,listings}

\definecolor{ablationPromptBody}{HTML}{F7F9FC}
\definecolor{ablationPromptTitle}{HTML}{E8EEF6}
\definecolor{ablationPromptBorder}{HTML}{B9C8DA}
\definecolor{ablationPromptInk}{HTML}{26364A}

\newtcblisting{agentprompt}[1]{
  enhanced,
  breakable,
  listing only,
  listing engine=listings,
  title={Prompt: \texttt{#1}},
  fonttitle=\small\bfseries,
  coltitle=black,
  colbacktitle=black!5,
  colback=black!5,
  colframe=black!70,
  boxrule=0.9pt,
  titlerule=0pt,
  arc=3pt,
  left=10pt, right=10pt, top=2pt, bottom=7pt,
  toptitle=6pt, bottomtitle=0pt,
  before skip=6pt, after skip=8pt,
  listing options={
    language={},
    basicstyle=\small\rmfamily,
    columns=fullflexible,
    keepspaces=true,
    breaklines=true,
    breakatwhitespace=true,
    breakautoindent=false,
    breakindent=0pt,
    showstringspaces=false,
    numbers=none,
    tabsize=2,
    aboveskip=0pt,
    belowskip=0pt
  }
}

\newtcblisting{ablationprompt}[1]{
  enhanced,
  breakable,
  listing only,
  listing engine=listings,
  title={Prompt: \texttt{#1}},
  fonttitle=\small\bfseries,
  coltitle=black,
  colbacktitle=black!5,
  colback=black!5,
  colframe=black!70,
  boxrule=0.9pt,
  titlerule=0pt,
  arc=3pt,
  left=10pt, right=10pt, top=2pt, bottom=7pt,
  toptitle=6pt, bottomtitle=0pt,
  before skip=8pt, after skip=10pt,
  listing options={
    language={},
    basicstyle=\small\rmfamily,
    columns=fullflexible,
    keepspaces=true,
    breaklines=true,
    breakatwhitespace=true,
    breakautoindent=false,
    breakindent=0pt,
    showstringspaces=false,
    numbers=none,
    tabsize=2,
    aboveskip=0pt,
    belowskip=0pt
  }
}

\newcommand{\appendixboxtitlefont}{\normalfont\rmfamily\bfseries\fontsize{9}{11}\selectfont}

\newtcolorbox{toolcard}[1]{
    enhanced, breakable,
    colback=white, colframe=black!50,
    colbacktitle=black!65, coltitle=white,
    fonttitle=\appendixboxtitlefont, fontupper=\small,
    title={#1}, boxrule=0.4pt, arc=0pt,
    left=7pt, right=7pt, top=5pt, bottom=5pt,
    toptitle=3pt, bottomtitle=3pt,
    before skip=8pt, after skip=8pt
  }
\newcommand{\toolfield}[1]{\par\smallskip\noindent\textbf{#1}\par\nobreak\noindent\ignorespaces}

\title{ReTAC: Residual Test-Time Agentic\\ Computation for Video Spatial Reasoning}

\author{%
Yike Wu$^{1}$ \quad Keyan Chen$^{1}$\thanks{Co-corresponding authors.} \quad Qixiong Wang$^{2}$ \quad Xi Xiang$^{2}$ \quad
Maofei Que$^{2}$ \\
Xiaolong Jiang$^{2}$ \quad Xiaoqin Zhang$^{3}$ \quad Ling Shao$^{4}$ \quad Shijian Lu$^{1}$\footnotemark[1]
\\[4pt]
$^{1}$Nanyang Technological University \qquad $^{2}$Xiaohongshu Inc. \\
$^{3}$Zhejiang University of Technology \qquad $^{4}$University of the Chinese Academy of Sciences \\
\href{https://yike-wu.github.io/ReTAC/}{\textcolor{magenta}{\texttt{https://yike-wu.github.io/ReTAC/}}}
}

\iclrfinalcopy
\begin{document}

\maketitle
\raggedbottom
\newlength{\retacSavedTextfloatsep}
\newlength{\retacSavedFloatsep}
\setlength{\retacSavedTextfloatsep}{\textfloatsep}
\setlength{\retacSavedFloatsep}{\floatsep}
\setlength{\textfloatsep}{8pt plus 2pt minus 2pt}
\setlength{\floatsep}{10pt plus 2pt minus 2pt}


\begin{abstract}
Test-time agentic computation can improve multimodal large language models (MLLMs) through planning, tool use, iterative perception, and reflection, but indiscriminate computation may duplicate resolved work, introduce noisy evidence, and perturb correct predictions. We propose that such computation should be \emph{residual}: allocate additional computation only to requirements unresolved by the base model and, after a localized failure, recompute only the affected states.
We instantiate this formulation in \textbf{ReTAC}, a training-free multiagent framework for video spatial reasoning. ReTAC organizes supplementary computation around unresolved requirements. It plans ordered steps to acquire missing spatial evidence and combines the observations with compatible native evidence for reasoning. In addition, it uses stage-aligned verification to localize failures to planning, perception, or reasoning. Dependency-based recovery recomputes only affected states while reusing unaffected ones.
Extensive experiments  show that ReTAC achieves the highest overall score among the compared methods across  three mainstream MLLMs and two widely adopted benchmarks. Further analyses show that gains from additional computation depend not only on how much is used, but also on where it is allocated and which valid intermediate states are preserved.
\end{abstract}

\section{Introduction}

Multimodal large language models (MLLMs) are increasingly augmented at test time with agentic computation, such as planning, external tools, iterative perception, and reflection, aiming to solve problems that native inference alone cannot resolve well. Video spatial reasoning makes the allocation of such computation particularly challenging because answer-relevant evidence is often distributed across frames and heterogeneous spatial cues. Solving a query often requires tracking object identities over time, accounting for camera motion, recovering depth and geometry, and aligning observations across spatial reference frames. These capabilities are relevant to embodied agents, AR/VR, and autonomous systems~\citep{yang2025embodiedbench,Dongfang_2026_CVPR,li2026ovo}. Existing methods improve access to spatial evidence through auxiliary representations, including depth maps, camera poses, cognitive maps, and structured 3D representations~\citep{qi2026gpt4scene,fan2026vlm,cheng2026grounded,hua2026unleashing,zhang2026think3d,ruan2026allospatial}, or through external tools that acquire additional evidence at inference time~\citep{zhang2026think3d,li2026skill,wang2026adatooler,dai2026s}. Additional evidence, however, is not necessarily beneficial: ViSRA reports that providing Qwen2.5-VL with a ground-truth cognitive map can reduce performance~\citep{mou2026visra}. This raises a more specific question: \emph{given what the base model has already resolved, what additional test-time computation is still needed?}

The answer depends on both the model and the specific instance. For the same video and question, one MLLM may resolve all required evidence natively, another may lack only a particular spatial relation, and a third may require additional perception. Adaptive agents condition tool use on task context, scene state, intermediate evidence, feedback, or acquired skills~\citep{li2026skill,wang2026adatooler}. Such adaptation does not explicitly distinguish requirements already supported by the current model from those that remain unresolved. Supplementary processing can therefore reacquire supported evidence, introduce noisy observations, and perturb an otherwise correct prediction. We formulate \emph{residual test-time agentic computation}: supplementary computation targets only requirements unresolved by native inference, while compatible native evidence is retained.

\begin{wrapfigure}{r}{0.685\linewidth}
\centering
\vspace{-2pt}
\includegraphics[width=\linewidth]{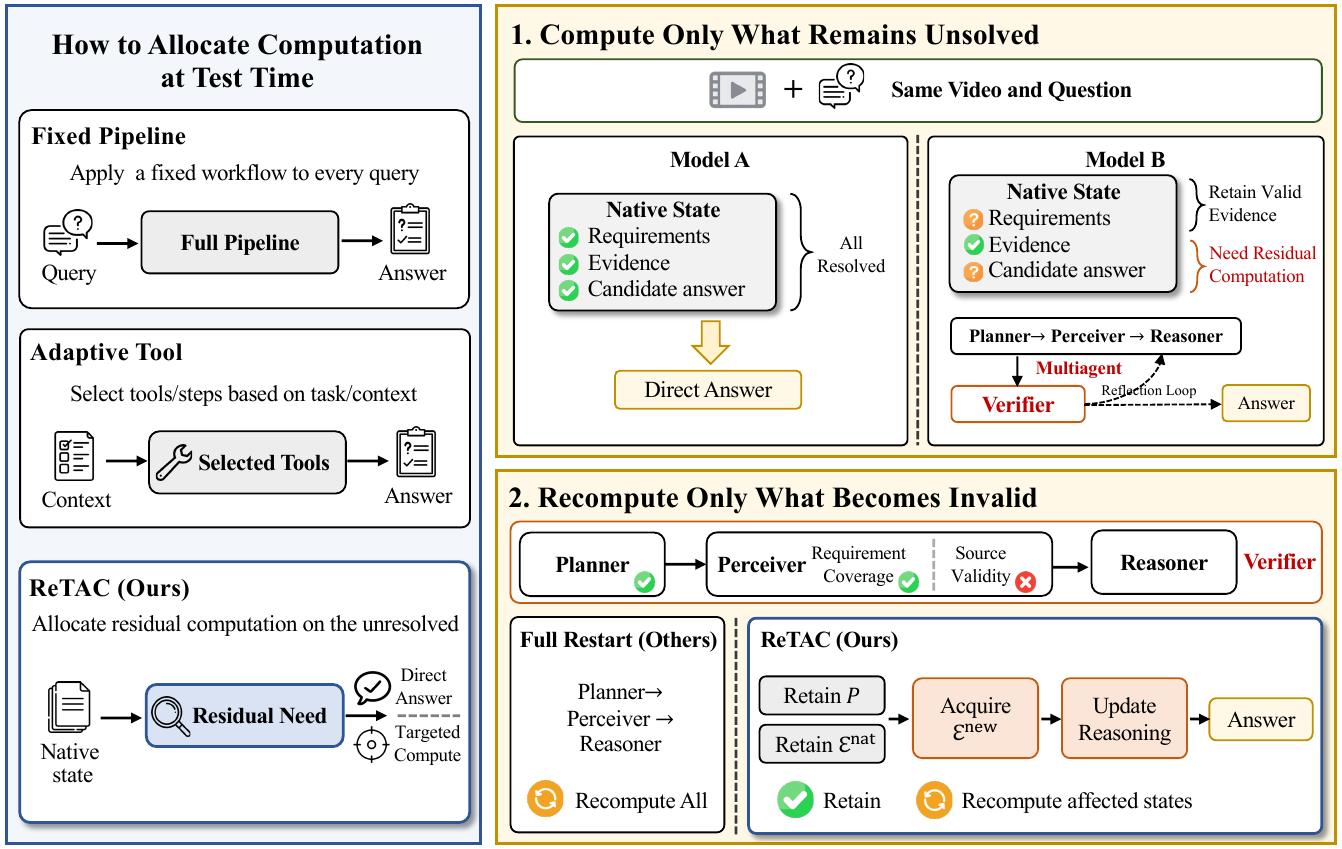}
\vspace{-0.5cm}
\caption{
\textbf{Residual test-time agentic computation.}
\textbf{Left:} fixed, context-adaptive, and residual computation allocation.
\textbf{Top right:} model--instance routing from the native state to direct return or targeted refinement.
\textbf{Bottom right:} dependency-aware selective recovery after a localized failure, while retaining valid evidence.
\vspace{-0.4cm}
}
\label{fig:motivation}
\end{wrapfigure}


Residual computation also applies to recovery. Agentic inference is multistage: planning may omit a required condition, perception may produce a source-inconsistent observation, and reasoning may misuse  valid evidence. Existing systems use reflection or consistency checks to assess evidence sufficiency, frame quality, or object consistency~\citep{mou2026visra,jiang2026refer}. However, when a failure is localized, restarting the entire pipeline discards valid intermediate state and repeats unaffected computation. Residual recovery instead recomputes only the invalidated state and its dependent states. Fig.~\ref{fig:motivation} summarizes these two principles: \emph{compute only what remains unresolved, and recompute only what becomes invalid}.

We instantiate these principles in \textbf{ReTAC} (\textbf{Re}sidual \textbf{T}est-Time \textbf{A}gentic \textbf{C}omputation), a training-free multiagent framework for video spatial reasoning. ReTAC first produces a structured native state containing task requirements, source-linked native evidence, unresolved requirements, and a candidate answer. If no unresolved requirement is reported, ReTAC returns the native candidate directly. Otherwise, residual planning constructs supplementary perception steps only for the unresolved requirements. Targeted spatial perception executes these steps, and the resulting evidence is integrated with native evidence for reasoning. ReTAC then applies stage-aligned verification to planning, perception, and reasoning. A localized failure triggers replanning, reperception, or rereasoning, while ReTAC preserves and reuses unaffected intermediate states. ReTAC therefore uses the current residual state to determine what additional computation and recovery are needed, rather than repeatedly executing a complete agentic pipeline.

We evaluate ReTAC on VSI-Bench-Debiased and STI-Bench with Qwen3-VL-8B, Gemini~3.1~Pro, and GPT-5.5. ReTAC achieves the highest overall score among the compared methods for all three models on both benchmarks. Controlled routing interventions directly test computation allocation: forcing supplementary processing on instances selected for direct return lowers performance, while bypassing supplementary processing on instances selected for residual refinement also degrades performance for all three models. At the default recovery budget, selective recovery uses fewer mean model calls than full restart for all three models, with higher overall scores for Qwen and GPT and only a slight 0.08-point decrease for Gemini. These results indicate that test-time scaling is allocation-sensitive: its utility depends not only on the amount of computation, but also on where computation is applied and which valid states are preserved. 

Our main contributions are threefold. First, we formulate \textbf{residual test-time agentic computation}, which allocates supplementary computation to requirements unresolved by the current model--instance pair and restricts recovery to invalidated states and their dependencies.
Second, we introduce \textbf{ReTAC}, a training-free multiagent framework that realizes this formulation through structured native-state generation, residual routing and planning, targeted spatial perception, stage-aligned verification, and dependency-based selective recovery.
Third, across three heterogeneous MLLMs and two video spatial reasoning benchmarks, ReTAC achieves the highest overall score among the compared methods in all six model--benchmark settings. Ablation studies and further analyses  support its two core designs: residual allocation targets computation to unresolved requirements, while  selective recovery reduces model calls relative to full restart with comparable overall performance.

\section{Related Work}

\paragraph{Video Spatial Reasoning with MLLMs.}
VSI-Bench and STI-Bench reveal persistent limitations of MLLMs in video spatial reasoning~\citep{yang2025thinking, li2025sti}. Existing methods mainly provide the model with richer spatial evidence through explicit scene representations or additional evidence gathering. Explicit representations take several forms: GPT4Scene uses bird's-eye views and consistent object markers to link global and local views~\citep{qi2026gpt4scene}, TRACE records scene context, camera trajectories, and object entities as structured text~\citep{hua2026unleashing}, and AlloSpatial combines geometric and semantic evidence in query-conditioned allocentric representations~\citep{ruan2026allospatial}. Additional spatial evidence can be obtained through Think3D's 3D reconstruction and iterative viewpoint changes~\citep{zhang2026think3d} or ViSRA's specialized spatial tools for multi-role inference~\citep{mou2026visra}. These approaches improve how spatial evidence is represented or acquired. ReTAC instead treats supplementary spatial processing as an on-demand mechanism, invoking it only when native evidence is insufficient.

\paragraph{Adaptive Test-Time Agentic Computation.}
Adaptive multimodal agents decide when and how to invoke tools during inference~\citep{wang2026adatooler,song2026adareasoner}. AdaTooler-V learns selective invocation from tool-benefit rewards~\citep{wang2026adatooler}, while AdaReasoner selects tools across multiple steps~\citep{song2026adareasoner}. DART recruits tools from agent disagreement, and ToolScope coordinates global planning with local visual perception~\citep{sivakumaran2026dart,deng2026toolscope}. Agent-X further evaluates stepwise reasoning and tool use in vision-centric tasks, revealing persistent failures in multimodal reasoning chains~\citep{ashraf2026agent}. For spatial reasoning, Skill-3D retrieves skills relevant to the scene and task for iterative evidence acquisition~\citep{li2026skill}. Their allocation signals include task context, intermediate outcomes, disagreement, and prior tool-use experience. ReTAC bases this decision on the residual state of the current model–instance pair, distinguishing supported native evidence from unresolved requirements.

\paragraph{Multimodal Reflection and Selective Recovery.}
Reflection-based methods revisit visual evidence or intermediate reasoning after an initial attempt. Recent methods encourage explicit self-reflection or evaluate intermediate reasoning steps with multimodal process
supervision~\citep{NEURIPS2025_2c84844a, wang2025visualprm}. VISCO studies visual critique and correction, showing that self-generated feedback can be unreliable~\citep{wu2025visco}. GThinker learns cue-guided visual rethinking~\citep{zhan2026gthinker}, while REVISOR returns to video segments relevant to the question~\citep{li2026revisor}. Refer-Agent uses reflection to refine frame selection and object analysis~\citep{jiang2026refer}, and ViSRA separates planning, reflection, execution, and answer synthesis across specialized roles~\citep{mou2026visra}. These methods improve evidence re-examination and feedback during reflection. ReTAC determines recovery scope after a localized failure: stage-aligned verification identifies whether the error lies in planning, perception, or reasoning, and dependency tracking limits recomputation to invalidated states while preserving unaffected states.

\section{Method}
\label{sec:method}

\subsection{Native State and Residual Routing}
\label{sec:overview}
\label{sec:native-assessment}

\begin{figure}[t]
\centering
\includegraphics[width=\linewidth]{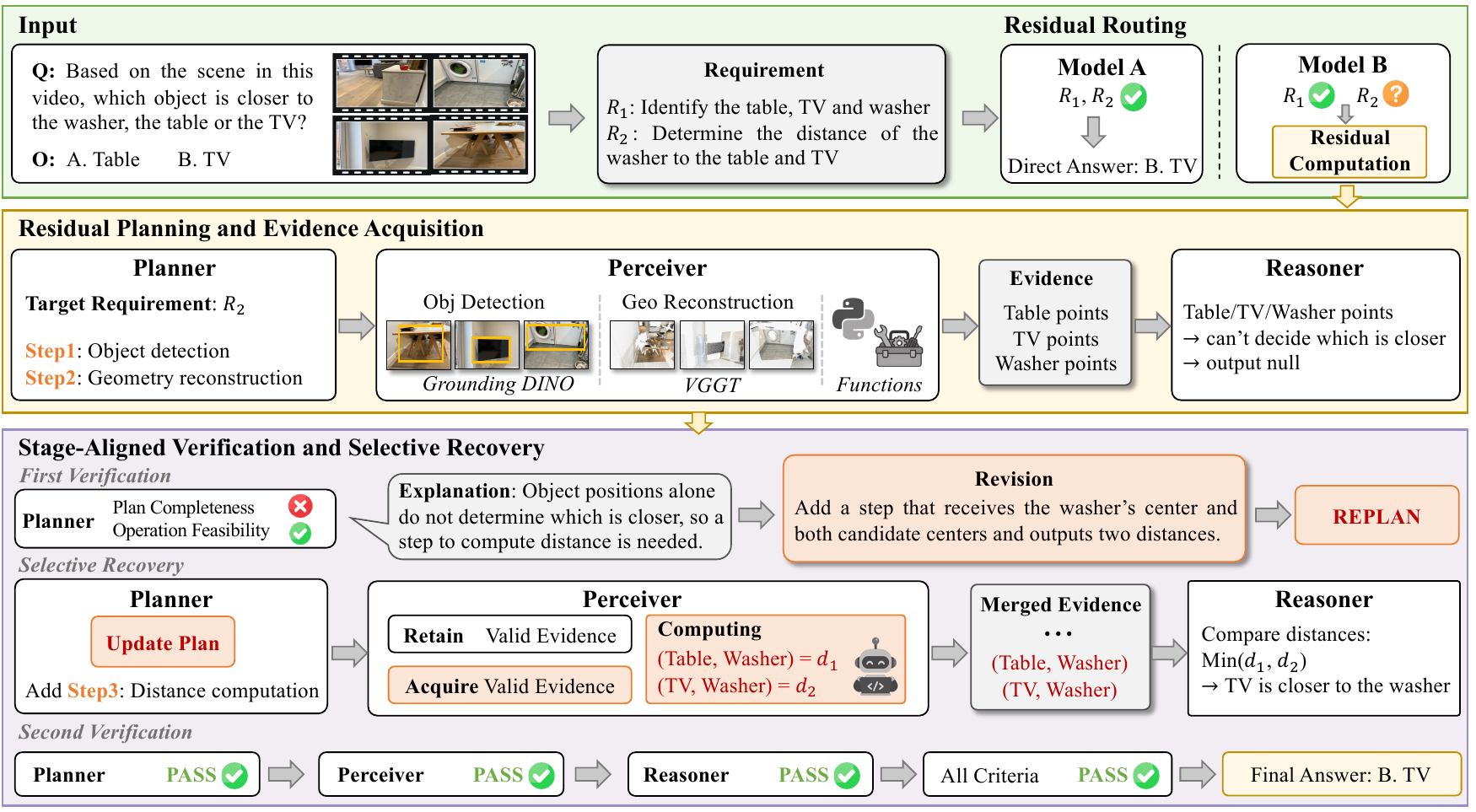}%
\vspace{-0.4cm}
\caption{\textbf{Overview of ReTAC.}
Native inference either returns the candidate when all identified requirements are supported or routes unresolved requirements to residual refinement. The illustrated refinement path detects an incomplete plan. \texttt{REPLAN} adds only the missing distance computation while retaining valid evidence.}
\label{fig:retac-overview}
\end{figure}

Given a video $V$, a spatial reasoning question $q$, and optional answer choices $\mathcal{O}$, ReTAC predicts an answer with a base MLLM $M$ and an operation interface $\mathcal{T}$. The same model serves as Planner, Perceiver, Reasoner, and Verifier under role-specific instructions, without task-specific training. ReTAC makes two allocation decisions: whether native inference requires supplementary computation and, after a localized verification failure, which states must be recomputed. Fig.~\ref{fig:retac-overview} gives an overview.

ReTAC begins with a single native call that receives only $x=(V,q,\mathcal{O})$, without tool descriptions or outputs, and produces the following state: 
\begin{equation}
S^{\mathrm{nat}}=\operatorname{Native}_M(x)
=\bigl(\mathcal{R}^{\mathrm{nat}},\mathcal{E}^{\mathrm{nat}},
\mathcal{U}^{\mathrm{nat}},\hat{y}^{\mathrm{nat}}\bigr),
\label{eq:overview-native}
\end{equation}
where $\mathcal{R}^{\mathrm{nat}}$ contains the identified task requirements, $\mathcal{E}^{\mathrm{nat}}$ contains source-linked native evidence, $\mathcal{U}^{\mathrm{nat}}$ contains unresolved requirements, and $\hat{y}^{\mathrm{nat}}$ is the native candidate answer. Each requirement is labeled \texttt{SUPPORTED} or \texttt{UNRESOLVED}, and supported requirements reference records in $\mathcal{E}^{\mathrm{nat}}$.

The requirements specify the information needed to answer the question, such as object identities, spatial relations, and temporal conditions. For example, native inference may identify the relevant objects but leave their spatial relation unresolved. This decomposition lets subsequent computation retain supported observations and acquire only the missing evidence. ReTAC routes the instance as
\begin{equation}
\hat{y}=
\begin{cases}
\hat{y}^{\mathrm{nat}}, & \mathcal{U}^{\mathrm{nat}}=\varnothing,\\
\operatorname{Refine}_M(x,S^{\mathrm{nat}},\mathcal{T}), & \text{otherwise},
\end{cases}
\label{eq:overview-routing}
\end{equation}
where $\operatorname{Refine}_M$ comprises residual planning, perception, reasoning, and verification with  selective recovery. The first branch returns \texttt{DIRECT} as a routing decision rather than a verification result.

\subsection{Residual Planning and Evidence Acquisition}
\label{sec:planning}

\paragraph{Residual Planning.}
For an instance routed to refinement, the Planner extends the native state only where additional information is needed. It constructs $P_0=(\mathcal{S}_0,\mathcal{R}_0,\mathcal{U}_0)$, where $\mathcal{S}_0$ is an ordered sequence of supplementary steps. Requirements omitted by native inference can be added as $\mathcal{R}^{\mathrm{add}}_0$:
\begin{equation}
\mathcal{R}_0=\mathcal{R}^{\mathrm{nat}}\cup\mathcal{R}^{\mathrm{add}}_0,
\qquad
\mathcal{U}_0=\mathcal{U}^{\mathrm{nat}}\cup\mathcal{R}^{\mathrm{add}}_0.
\label{eq:residual-requirements}
\end{equation}

Each step targets a requirement and invokes either a single primitive operation or a composition of primitives. For each primitive operation, the plan specifies the arguments, input references, dependencies, and expected output. The Planner constructs each composition according to the requirements of the task. Inputs may reference the video, native evidence, or preceding operation outputs.
A step is included only if it resolves a requirement in $\mathcal{U}_0$ or supplies a prerequisite for doing so. The interface $\mathcal{T}$ provides GroundingDINO for object detection~\citep{liu2024grounding}, VGGT for geometry estimation~\citep{wang2025vggt}, and predefined spatial functions. Appendix~\ref{app:tool_definitions} describes these tools and functions in detail.

\paragraph{Targeted Perception.}
The Perceiver executes $\mathcal{S}_0$ in dependency order using the declared operations and arguments. Each observation is stored as:
\begin{equation}
e_j=(r_j,o_j,\sigma_j,p_j,\mathcal{D}_j),
\label{eq:evidence}
\end{equation}
where $r_j$ is the target requirement, $o_j$ the observed content, $\sigma_j$ the sources supporting that content, $p_j$ the step producing the observation ($p_j=\mathrm{nat}$ for native observations), and $\mathcal{D}_j$ the input on which the observation depends. Each record  preserves the object identity, temporal reference, coordinate frame, and scale required to interpret $o_j$.  Source references $\sigma_j$ point to sampled frames or raw tool outputs, allowing later verification to distinguish errors in evidence acquisition from errors in the Perceiver's interpretation. Failed acquisitions are logged without creating an observation. 

Supplementary observations $\mathcal{E}^{\mathrm{new}}$ are merged with $\mathcal{E}^{\mathrm{nat}}$ to form the active evidence set $\mathcal{E}_0$  for reasoning. Conflicting observations remain in $\mathcal{E}_0$ with  provenance and dependencies for verification. Raw outputs are retained in the execution history independently of observations derived from them. 


\paragraph{Evidence-Grounded Reasoning.}
The Reasoner computes
$(Z_0, \hat{y}_0) = \operatorname{Reason}_M(q, O, P_0, \mathcal{E}_0)$,
where $Z_0$ is the reasoning trace. Each critical claim must reference
active evidence or a preceding supported claim. The Reasoner introduces
no new visual observations. The resulting state
$\Sigma_0 = (P_0, \mathcal{E}_0, Z_0, \hat{y}_0)$ remains provisional
until verification.

\subsection{Stage-Aligned Verification}
\label{sec:verification-control}

\begin{table}[t]
\centering
\setlength{\abovecaptionskip}{0pt}
\setlength{\belowcaptionskip}{4pt}
\caption{\textbf{VSI-Bench-Debiased results.} Parentheses report ReTAC minus the corresponding method within the same base model. Count, Abs.D, Obj.S, Room, Rel.Dir, Rel.D, Order, and Route denote object counting, absolute distance, object size, room size, relative direction, relative distance, appearance order, and route planning. S\&T denotes See\&Trek~\citep{li2025seetrek}.}
\label{tab:exp-main-vsi}
\retacTableSetup
\resizebox{\linewidth}{!}{%
\begin{tabular}{@{}
>{\centering\arraybackslash}m{\retacBaseWidth}
>{\centering\arraybackslash}m{\retacMethodWidth}
*{9}{>{\centering\arraybackslash}m{\retacMetricWidth}}
@{}}
\toprule
\multirow{2}{*}{\textbf{Base}} &
\multirow{2}{*}{\textbf{Method}} &
\multicolumn{4}{c}{\textbf{Numerical (MRA, \%)}} &
\multicolumn{4}{c}{\textbf{Multiple Choice (Acc., \%)}} &
\multirow{2}{*}{\textbf{Avg.}} \\
\cmidrule(lr){3-6}\cmidrule(lr){7-10}
& & Count & Abs.D & Obj.S & Room & Rel.Dir & Rel.D & Order & Route & \\
\midrule
\multirow[c]{5}{*}{\shortstack{Qwen3\\VL 8B}} & Direct
& 44.82\retacUp{8.37} & 39.75\retacUp{3.80} & 59.24\retacUp{2.37} & 58.60\retacDown{1.30} & 41.67\retacUp{13.59} & 49.35\retacUp{5.49} & 60.38\retacUp{8.49} & 28.07\retacUp{5.26} & 47.73\retacUp{5.76} \\
 & CoT
& 45.12\retacUp{8.07} & 39.63\retacUp{3.92} & 59.15\retacUp{2.46} & 58.32\retacDown{1.02} & 43.23\retacUp{12.03} & 50.96\retacUp{3.88} & 61.95\retacUp{6.92} & 30.41\retacUp{2.92} & 48.59\retacUp{4.90} \\
 & S\&T
& 43.55\retacUp{9.64} & 39.98\retacUp{3.57} & \textbf{62.04}\retacDown{0.43} & 55.55\retacUp{1.75} & 37.20\retacUp{18.06} & \textbf{55.16}\retacDown{0.32} & 66.35\retacUp{2.52} & 26.32\retacUp{7.01} & 48.27\retacUp{5.22} \\
 & Think3D
& 44.86\retacUp{8.33} & 39.82\retacUp{3.73} & 60.14\retacUp{1.47} & \textbf{58.75}\retacDown{1.45} & 42.56\retacUp{12.70} & 52.37\retacUp{2.47} & 61.85\retacUp{7.02} & 31.58\retacUp{1.75} & 48.99\retacUp{4.50} \\
 & \retacCell{\textbf{ReTAC}}
& \retacCell{\textbf{53.19}} & \retacCell{\textbf{43.55}} & \retacCell{61.61} & \retacCell{57.30} & \retacCell{\textbf{55.26}} & \retacCell{54.84} & \retacCell{\textbf{68.87}} & \retacCell{\textbf{33.33}} & \retacCell{\textbf{53.49}} \\

\midrule
\multirow[c]{5}{*}{\shortstack{Gemini\\3.1 Pro}} & Direct
& 41.12\retacUp{7.76} & 31.31\retacUp{3.62} & 56.26\retacUp{2.38} & 46.20\retacUp{6.05} & 46.44\retacUp{7.35} & 59.03\retacUp{7.10} & 64.15\retacUp{2.52} & 45.61\retacUp{3.51} & 48.77\retacUp{5.03} \\
 & CoT
& 42.36\retacUp{6.52} & 31.57\retacUp{3.36} & 57.12\retacUp{1.52} & 45.98\retacUp{6.27} & 48.96\retacUp{4.83} & 62.15\retacUp{3.98} & 66.46\retacUp{0.21} & 47.37\retacUp{1.75} & 50.24\retacUp{3.56} \\
 & S\&T
& 44.26\retacUp{4.62} & 30.20\retacUp{4.73} & \textbf{59.49}\retacDown{0.85} & 44.10\retacUp{8.15} & 50.44\retacUp{3.35} & \textbf{68.39}\retacDown{2.26} & \textbf{69.70}\retacDown{3.03} & 48.53\retacUp{0.59} & 51.89\retacUp{1.91} \\
 & Think3D
& 43.37\retacUp{5.51} & 30.32\retacUp{4.61} & 57.23\retacUp{1.41} & 47.33\retacUp{4.92} & 52.65\retacUp{1.14} & 63.76\retacUp{2.37} & 68.24\retacDown{1.57} & 47.37\retacUp{1.75} & 51.28\retacUp{2.52} \\
 & \retacCell{\textbf{ReTAC}}
& \retacCell{\textbf{48.88}} & \retacCell{\textbf{34.93}} & \retacCell{58.64} & \retacCell{\textbf{52.25}} & \retacCell{\textbf{53.79}} & \retacCell{66.13} & \retacCell{66.67} & \retacCell{\textbf{49.12}} & \retacCell{\textbf{53.80}} \\

\midrule
\multirow[c]{5}{*}{\shortstack{GPT-5.5}} & Direct
& 49.56\retacUp{0.40} & \textbf{45.51}\retacDown{0.58} & 63.88\retacUp{1.83} & 57.55\retacUp{0.85} & 64.92\retacUp{4.64} & 67.42\retacUp{0.75} & 71.70\retacUp{1.88} & 67.54\retacUp{1.18} & 61.01\retacUp{1.37} \\
 & CoT
& 50.23\retacDown{0.27} & 44.57\retacUp{0.36} & 63.82\retacUp{1.89} & 57.87\retacUp{0.53} & 65.66\retacUp{3.90} & 67.96\retacUp{0.21} & 72.64\retacUp{0.94} & 68.42\retacUp{0.30} & 61.40\retacUp{0.98} \\
 & S\&T
& \textbf{50.60}\retacDown{0.64} & 45.07\retacDown{0.14} & 62.63\retacUp{3.08} & 53.00\retacUp{5.40} & 66.69\retacUp{2.87} & 65.27\retacUp{2.90} & 70.75\retacUp{2.83} & 65.79\retacUp{2.93} & 59.98\retacUp{2.40} \\
 & Think3D
& 48.79\retacUp{1.17} & 43.28\retacUp{1.65} & 63.77\retacUp{1.94} & 57.73\retacUp{0.67} & 67.45\retacUp{2.11} & \textbf{68.71}\retacDown{0.54} & 72.43\retacUp{1.15} & 67.84\retacUp{0.88} & 61.26\retacUp{1.12} \\
 & \retacCell{\textbf{ReTAC}}
& \retacCell{49.96} & \retacCell{44.93} & \retacCell{\textbf{65.71}} & \retacCell{\textbf{58.40}} & \retacCell{\textbf{69.56}} & \retacCell{68.17} & \retacCell{\textbf{73.58}} & \retacCell{\textbf{68.72}} & \retacCell{\textbf{62.38}} \\

\bottomrule
\end{tabular}%
}
\end{table}
To localize errors before recovery, the Verifier checks the current state $\Sigma_t=(P_t,\mathcal{E}_t,Z_t,\hat{y}_t)$ in the order of planning, perception, and reasoning, where $t$ denotes the number of completed recovery actions and $t=0$ the initial refinement state. The base model evaluates semantic support, while deterministic checks validate interfaces, references, and traceable computations.

\paragraph{Planning.}
Given $(q,\mathcal{O},P_t,\mathcal{T})$, the Verifier checks \emph{plan completeness} and \emph{operation feasibility}: whether the plan includes  steps needed to address every requirement not supported by native evidence and whether declared operations can produce the required evidence. To isolate planning errors, this stage does not access the video, observations, reasoning trace, or candidate answer. An omission is considered necessary if the missing information could change the answer while all represented requirements are held fixed. Feasibility is assessed from  documented operation interfaces.

\paragraph{Perception.}
The Verifier checks \emph{source validity} and \emph{requirement coverage} against the cited frames and raw outputs. Native and supplementary observations follow the same criteria: each active record must be supported by its source, and requirements in $\mathcal{R}_t$ must be resolved by valid active evidence.

\paragraph{Reasoning.}
The Verifier checks \emph{evidence support}, \emph{inference validity}, and \emph{answer consistency} along $\mathcal{E}_t\rightarrow Z_t\rightarrow\hat{y}_t$. These checks validate the premises, the resulting semantic or computational derivations, and the mapping from the final conclusion to the returned answer, including option mapping or numerical normalization when required.

Each criterion returns a verdict $v_{t,c}\in\{\texttt{PASS},\texttt{UNKNOWN},\texttt{FAIL}\}$, with a target and justification. For stage $a\in\{\mathrm{plan},\mathrm{perc},\mathrm{reason}\}$ with criteria $\mathcal{C}_a$, the stage verdict is:
\begin{equation}
v_t^a=\min_{c\in\mathcal{C}_a}v_{t,c},
\qquad
\texttt{FAIL}\prec\texttt{UNKNOWN}\prec\texttt{PASS}.
\label{eq:verification-aggregation}
\end{equation}
The stage passes only if all criteria pass. A single \texttt{FAIL} makes the stage fail and any \texttt{UNKNOWN} makes it non-passing. Verification stops at the first non-passing stage and returns a localized diagnosis with a revision target. If a later check raises a concrete concern about an earlier pass, only the implicated upstream criterion is rechecked before recovery is selected. Appendix~\ref{app:verification-details} specifies the criteria and stage-specific inputs.

\begin{table}[t]
\centering
\setlength{\abovecaptionskip}{0pt}
\setlength{\belowcaptionskip}{4pt}
\caption{\textbf{STI-Bench results.} Parentheses report ReTAC minus the corresponding method within the same base model. DM, SR, VG, DPL, SA, EO, TD, and PE denote dimensional measurement, spatial relation, 3D video grounding, displacement and path length, speed and acceleration, egocentric orientation, trajectory description, and pose estimation. S\&T denotes See\&Trek~\citep{li2025seetrek}.}
\label{tab:exp-main-sti}
\retacTableSetup
\resizebox{\linewidth}{!}{%
\begin{tabular}{@{}
>{\centering\arraybackslash}m{\retacBaseWidth}
>{\centering\arraybackslash}m{\retacMethodWidth}
*{9}{>{\centering\arraybackslash}m{\retacMetricWidth}}
@{}}
\toprule
\multirow{2}{*}{\textbf{Base}} &
\multirow{2}{*}{\textbf{Method}} &
\multicolumn{3}{c}{\textbf{Static Understanding}} &
\multicolumn{5}{c}{\textbf{Dynamic Understanding}} &
\multirow{2}{*}{\textbf{Avg.}} \\
\cmidrule(lr){3-5}\cmidrule(lr){6-10}
& & DM & SR & VG & DPL & SA & EO & TD & PE & \\
\midrule
\multirow[c]{5}{*}{\shortstack{Qwen3\\VL 8B}} & Direct
& 29.76\retacDown{0.23} & 47.26\retacUp{4.79} & 39.75\retacUp{4.73} & 19.83\retacUp{9.50} & 38.37\retacUp{3.62} & 57.84\retacUp{5.94} & 39.74\retacUp{20.52} & 57.22\retacUp{0.28} & 39.87\retacUp{4.63} \\
 & CoT
& 30.10\retacDown{0.57} & 48.63\retacUp{3.42} & 40.06\retacUp{4.42} & 23.46\retacUp{5.87} & 39.88\retacUp{2.11} & 58.92\retacUp{4.86} & 46.15\retacUp{14.11} & 56.94\retacUp{0.56} & 41.23\retacUp{3.27} \\
 & S\&T
& 29.76\retacDown{0.23} & 47.03\retacUp{5.02} & 37.85\retacUp{6.63} & 23.18\retacUp{6.15} & 37.16\retacUp{4.83} & 62.16\retacUp{1.62} & 44.45\retacUp{15.81} & 55.83\retacUp{1.67} & 40.28\retacUp{4.22} \\
 & Think3D
& \textbf{34.95}\retacDown{5.42} & 50.00\retacUp{2.05} & 38.17\retacUp{6.31} & 28.59\retacUp{0.74} & 35.04\retacUp{6.95} & 56.76\retacUp{7.02} & 42.31\retacUp{17.95} & 53.42\retacUp{4.08} & 40.87\retacUp{3.63} \\
 & \retacCell{\textbf{ReTAC}}
& \retacCell{29.53} & \retacCell{\textbf{52.05}} & \retacCell{\textbf{44.48}} & \retacCell{\textbf{29.33}} & \retacCell{\textbf{41.99}} & \retacCell{\textbf{63.78}} & \retacCell{\textbf{60.26}} & \retacCell{\textbf{57.50}} & \retacCell{\textbf{44.50}} \\

\midrule
\multirow[c]{5}{*}{\shortstack{Gemini\\3.1 Pro}} & Direct
& 40.95\retacUp{2.65} & 61.64\retacDown{0.68} & 43.11\retacUp{0.11} & 33.24\retacUp{4.47} & 41.79\retacUp{2.42} & 63.06\retacUp{0.18} & 61.11\retacUp{8.12} & 70.65\retacUp{0.18} & 49.47\retacUp{1.86} \\
 & CoT
& 40.48\retacUp{3.12} & \textbf{62.33}\retacDown{1.37} & 42.90\retacUp{0.32} & 34.92\retacUp{2.79} & 42.60\retacUp{1.61} & 63.78\retacDown{0.54} & 65.38\retacUp{3.85} & 70.28\retacUp{0.55} & 50.00\retacUp{1.33} \\
 & S\&T
& 40.60\retacUp{3.00} & 59.81\retacUp{1.15} & 40.59\retacUp{2.63} & 34.82\retacUp{2.89} & 42.90\retacUp{1.31} & \textbf{66.85}\retacDown{3.61} & 59.40\retacUp{9.83} & 69.17\retacUp{1.66} & 49.37\retacUp{1.96} \\
 & Think3D
& 42.56\retacUp{1.04} & 62.10\retacDown{1.14} & 40.80\retacUp{2.42} & 36.59\retacUp{1.12} & 42.90\retacUp{1.31} & 62.35\retacUp{0.89} & 67.95\retacUp{1.28} & 69.26\retacUp{1.57} & 50.08\retacUp{1.25} \\
 & \retacCell{\textbf{ReTAC}}
& \retacCell{\textbf{43.60}} & \retacCell{60.96} & \retacCell{\textbf{43.22}} & \retacCell{\textbf{37.71}} & \retacCell{\textbf{44.21}} & \retacCell{63.24} & \retacCell{\textbf{69.23}} & \retacCell{\textbf{70.83}} & \retacCell{\textbf{51.33}} \\

\midrule
\multirow[c]{5}{*}{\shortstack{GPT-5.5}} & Direct
& 36.33\retacUp{5.19} & 69.18\retacDown{1.14} & 41.64\retacUp{3.15} & 37.71\retacUp{6.98} & 45.32\retacUp{1.31} & 51.35\retacDown{0.36} & 69.23\retacUp{1.71} & 70.00\retacDown{0.37} & 49.61\retacUp{2.53} \\
 & CoT
& 37.02\retacUp{4.50} & \textbf{69.86}\retacDown{1.82} & 41.32\retacUp{3.47} & 39.66\retacUp{5.03} & 46.22\retacUp{0.41} & 51.89\retacDown{0.90} & 70.51\retacUp{0.43} & 69.72\retacDown{0.09} & 50.24\retacUp{1.90} \\
 & S\&T
& 37.60\retacUp{3.92} & 65.75\retacUp{2.29} & 39.96\retacUp{4.83} & 38.55\retacUp{6.14} & \textbf{47.63}\retacDown{1.00} & \textbf{53.15}\retacDown{2.16} & \textbf{71.36}\retacDown{0.42} & \textbf{70.19}\retacDown{0.56} & 50.08\retacUp{2.06} \\
 & Think3D
& 39.67\retacUp{1.85} & 68.72\retacDown{0.68} & 39.86\retacUp{4.93} & 42.83\retacUp{1.86} & 46.32\retacUp{0.31} & 50.45\retacUp{0.54} & 70.51\retacUp{0.43} & 69.82\retacDown{0.19} & 50.76\retacUp{1.38} \\
 & \retacCell{\textbf{ReTAC}}
& \retacCell{\textbf{41.52}} & \retacCell{68.04} & \retacCell{\textbf{44.79}} & \retacCell{\textbf{44.69}} & \retacCell{46.63} & \retacCell{50.99} & \retacCell{70.94} & \retacCell{69.63} & \retacCell{\textbf{52.14}} \\

\bottomrule
\end{tabular}%
}
\end{table}

\subsection{Selective Recovery}
\label{sec:rollback}

\paragraph{Recovery Scope.}
The controller selects a recovery action only when verification supports a concrete revision. \texttt{REPLAN} revises requirements or acquisition steps, including operations, arguments, inputs, or dependencies. \texttt{REPERCEIVE} repeats an affected acquisition under the unchanged plan. If the acquisition procedure itself must change, the controller instead selects \texttt{REPLAN}. \texttt{REREASON} preserves the plan and evidence and regenerates the derivation and answer. Each recovery call receives the correction target, diagnosis, proposed revision, and accumulated recovery history.

\paragraph{Dependency-Based Reuse.}
\label{sec:state-reuse}
Let $\Delta_t$ denote the records or steps directly changed or invalidated by recovery. The affected set is
$\mathcal{I}_t=\Delta_t\cup\operatorname{Desc}(\Delta_t)$,
where $\operatorname{Desc}$ contains all transitive dependents under the recorded input and support references. Replanning retains evidence compatible with the revised plan, reperception preserves unrelated evidence, and rereasoning reuses the plan and all active evidence. Superseded records remain in the execution history but are removed from the active evidence set. Cached checks are reused only when their supporting records and justifications remain valid. Replacing one observation, for example, invalidates its source check and any coverage or reasoning checks that depend on it, while checks on unrelated observations can be retained. Every recovery regenerates $(Z_{t+1},\hat{y}_{t+1})$ as a single unit, while unaffected plan steps, evidence, and verification checks are reused.

\paragraph{Reverification and Termination.}
\label{sec:termination}
The revised state is verified after each recovery. ReTAC returns \texttt{VERIFIED} only when planning, perception, and reasoning all pass; the status denotes acceptance under the specified criteria rather than ground-truth correctness. The budget $T$ counts recovery actions, with verification performed at $t=0$ and after every permitted action, including the last. If no supported revision is available or the budget is exhausted before acceptance, ReTAC returns the current answer with status \texttt{UNVERIFIED}. Algorithm~\ref{alg:retac} in Appendix~\ref{app:retac-procedure} gives the full procedure.

\section{Experiments}
\label{sec:exp}

\begin{table}[t]
\centering
\caption{\textbf{Component replacements.} 
Parentheses report each variant minus ReTAC in percentage points.
Gen. Plan. and Gen. Refl. denote General Planning and General Reflection,
respectively.}
\label{tab:ablation}
\retacTableSetup
\resizebox{\linewidth}{!}{%
\begin{tabular}{@{}
>{\centering\arraybackslash}m{\retacBaseWidth}
>{\centering\arraybackslash}m{\retacMethodWidth}
*{9}{>{\centering\arraybackslash}m{\retacMetricWidth}}
@{}}
\toprule
\multirow{2}{*}{\textbf{Base}} &
\multirow{2}{*}{\textbf{Method}} &
\multicolumn{4}{c}{\textbf{Numerical (MRA, \%)}} &
\multicolumn{4}{c}{\textbf{Multiple Choice (Acc., \%)}} &
\multirow{2}{*}{\textbf{Avg.}} \\
\cmidrule(lr){3-6}\cmidrule(lr){7-10}
& & Count & Abs.D & Obj.S & Room & Rel.Dir & Rel.D & Order & Route & \\
\midrule
\multirow[c]{3}{*}{\shortstack{Qwen3\\VL 8B}} & \retacCell{\textbf{ReTAC}}
& \retacCell{\textbf{53.19}} & \retacCell{\textbf{43.55}} & \retacCell{\textbf{61.61}} & \retacCell{57.30} & \retacCell{\textbf{55.26}} & \retacCell{\textbf{54.84}} & \retacCell{\textbf{68.87}} & \retacCell{\textbf{33.33}} & \retacCell{\textbf{53.49}} \\
 & \mbox{Gen. Plan.}
& 51.69\retacDown{1.50} & 41.16\retacDown{2.39} & 60.89\retacDown{0.72} & \textbf{57.92}\retacUp{0.62} & 52.58\retacDown{2.68} & 49.78\retacDown{5.06} & 65.41\retacDown{3.46} & 31.58\retacDown{1.75} & 51.37\textbf{\retacDown{2.12}} \\
 & \mbox{Gen. Refl.}
& 52.43\retacDown{0.76} & 42.39\retacDown{1.16} & 61.54\retacDown{0.07} & 56.13\retacDown{1.17} & 53.23\retacDown{2.03} & 52.36\retacDown{2.48} & 64.78\retacDown{4.09} & 29.82\retacDown{3.51} & 51.58\textbf{\retacDown{1.91}} \\
\midrule
\multirow[c]{3}{*}{\shortstack{GPT-5.5}} & \retacCell{\textbf{ReTAC}}
& \retacCell{\textbf{49.96}} & \retacCell{\textbf{44.93}} & \retacCell{\textbf{65.71}} & \retacCell{\textbf{58.40}} & \retacCell{\textbf{69.56}} & \retacCell{\textbf{68.17}} & \retacCell{\textbf{73.58}} & \retacCell{\textbf{68.72}} & \retacCell{\textbf{62.38}} \\
 & \mbox{Gen. Plan.}
& 49.53\retacDown{0.43} & 44.24\retacDown{0.69} & 64.13\retacDown{1.58} & 58.20\retacDown{0.20} & 67.92\retacDown{1.64} & 67.10\retacDown{1.07} & 72.33\retacDown{1.25} & 66.67\retacDown{2.05} & 61.27\textbf{\retacDown{1.11}} \\
 & \mbox{Gen. Refl.}
& 49.18\retacDown{0.78} & 44.21\retacDown{0.72} & 65.38\retacDown{0.33} & 57.03\retacDown{1.37} & 68.27\retacDown{1.29} & 65.81\retacDown{2.36} & 72.01\retacDown{1.57} & 64.91\retacDown{3.81} & 60.85\textbf{\retacDown{1.53}} \\
\bottomrule
\end{tabular}%
}
\end{table}

\subsection{Experimental Setup}
\label{sec:setup}

We evaluate ReTAC with Qwen3-VL-8B~\citep{bai2025qwen3}, Gemini~3.1~Pro, and GPT-5.5 on VSI-Bench-Debiased~\citep{yang2025thinking, brown2025benchmarkdesignerstraintest} and STI-Bench~\citep{li2025sti}. VSI-Bench-Debiased uses mean relative accuracy (MRA) for numerical tasks and accuracy for multiple-choice tasks, averaging eight task scores uniformly. STI-Bench uses accuracy for all eight tasks and a sample-weighted overall score. We report means over three runs for the main experiments, component
  ablations, and recovery cost and budget analyses. The routing and
  answer-transition analyses use a single run for each base model. We uniformly sample 32 frames per video and set the default recovery budget $T=2$. All tool-enabled variants share the same operation interface. The same base MLLM instantiates the Planner, Perceiver, Reasoner, and Verifier under role-specific instructions shared across models.  Appendix~\ref{app:exp-details} provides  full evaluation and intervention protocols.

\subsection{Main Results}
\label{sec:main}

We compare ReTAC with Direct QA, chain-of-thought prompting (CoT), See\&Trek~\citep{li2025seetrek}, and Think3D~\citep{zhang2026think3d}. ReTAC obtains the highest score among the compared methods in all six model--benchmark settings (Tables~\ref{tab:exp-main-vsi} and~\ref{tab:exp-main-sti}). 

On VSI-Bench-Debiased, the clearest gains occur on tasks that require spatial evidence across frames or viewpoints. Relative direction improves for all three models, while Qwen and Gemini also improve on counting and appearance order. These categories align with the object localization, temporal evidence, and observer-frame computation targeted by residual refinement. Absolute distance and room size are less consistent across models, which may partly reflect VGGT's lack of metric scale. This task-wise pattern links the gains to the evidence available through the tools.

On STI-Bench, scores for displacement and path length improve substantially for all three base models, while trajectory description shows the largest gains for Qwen and Gemini. Both tasks require spatial evidence to be accumulated over time, matching ReTAC's targeted acquisition of unresolved temporal and geometric cues. The gains vary across tasks and models: several categories regress slightly, and GPT gains less than Gemini on VSI-Bench-Debiased but more on STI-Bench. This reversal suggests that the benefit of residual refinement depends more on the unresolved requirements of a model--instance pair and the evidence exposed by the tools than on base model strength alone.

\subsection{Residual Computation Allocation}
\label{sec:ablation}

All subsequent analyses use VSI-Bench-Debiased. Component replacements are evaluated with Qwen and GPT, while routing and recovery analyses use all three models.

\paragraph{Native-State Conditioning.}
To test the role of native-state conditioning, we replace residual routing and planning with a general adaptive planner that receives the input, task requirements, and tool interfaces, but not native support labels or native evidence. It can return the stored native answer or acquire evidence for the full task. Performance decreases for both models, with lower scores in 15 of 16 task-level comparisons (Table~\ref{tab:ablation}). These results indicate that, under a limited budget, focusing computation on the unresolved is more effective than reacquiring evidence for the full task.

\begin{figure}[t]
\centering
\begin{minipage}[t]{0.48\textwidth}
\vspace{0pt}
\centering
\includegraphics[width=\linewidth,height=4.2cm,keepaspectratio]{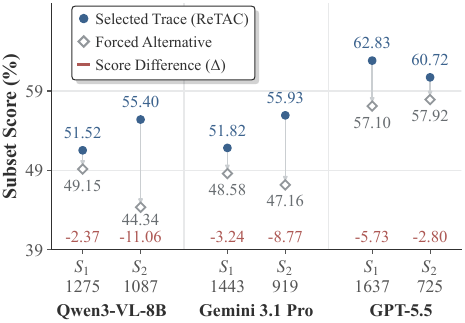}
\vspace{-0.4cm}
\caption{\textbf{Routing interventions on fixed subsets.} $S_1$ and $S_2$ are direct-return and refinement subsets. Circles show selected-route scores, diamonds the forced alternative, and $\Delta$ labels the within-subset change.}
\label{fig:exp-routing}
\end{minipage}\hfill
\begin{minipage}[t]{0.48\textwidth}
\vspace{0pt}
\centering
\includegraphics[width=\linewidth,height=4.2cm,keepaspectratio]{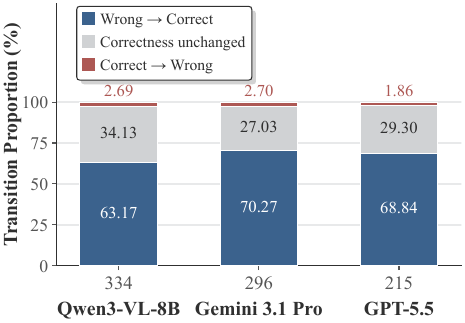}
\vspace{-0.4cm}
\caption{\textbf{Answer transitions after recovery.} Bars partition recovered instances into wrong-to-correct, unchanged correctness, and correct-to-wrong. X-axis labels show the number of recovered instances.}
\label{fig:reflection-outcomes}
\end{minipage}\hfill
\par\vspace{-4pt}
\end{figure}

\paragraph{Routing Interventions.}
For each model, we keep ReTAC's original direct return ($S_1$) and residual refinement ($S_2$) subsets fixed. On $S_1$, we force supplementary acquisition for every native requirement while retaining native evidence. On $S_2$, we suppress refinement and return the native candidate. Subset scores are computed by averaging per-question scores within each fixed routing subset. Both forced alternatives reduce subset performance for every model (Fig.~\ref{fig:exp-routing}). The larger penalty changes across base models: Qwen and Gemini lose more performance when refinement is suppressed, whereas GPT suffers a larger drop when supplementary acquisition is forced. This paired result supports model--instance-specific computation allocation rather than a uniformly deeper inference.

\newlength{\retacSavedAnalysisParskip}
\setlength{\retacSavedAnalysisParskip}{\parskip}
\setlength{\parskip}{4pt}
\subsection{Residual Recovery}
\label{sec:reflection-analysis}

\begingroup
\setlength{\intextsep}{6pt}
\begin{wrapfigure}{r}{0.6\linewidth}
\centering
\includegraphics[width=\linewidth]{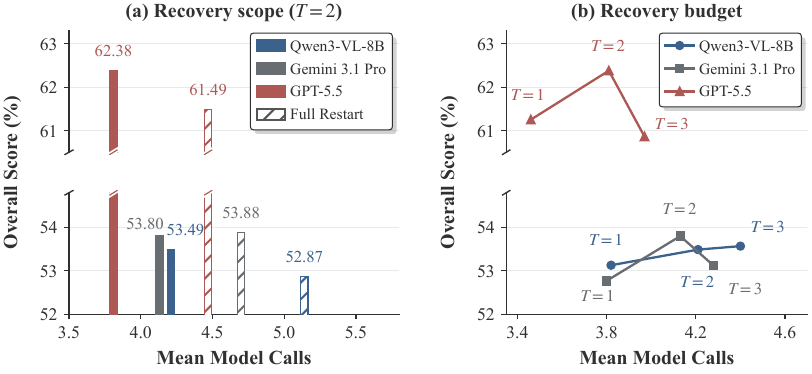}
\vspace{-10pt}
\setlength{\abovecaptionskip}{4pt}
\caption{
\textbf{Recovery scope and budget.}
  (a) Selective recovery versus full restart  at $T=2$.
  (b) Score--cost trade-offs as $T$ increases.
  Beyond $T=2$, calls increase with little or negative score gains.
}
\label{fig:correction-efficiency}
\vspace{-\baselineskip}
\end{wrapfigure}

\paragraph{Structured Verification and Localized Revision.}
Replacing stage-aligned verification and selective recovery with general
reflection and full restart lowers every task score for both evaluated
models (Table~\ref{tab:ablation}). The reflector receives the same trajectory and sources, so the comparison concerns how errors are localized and revisions are organized. The degradation suggests that effective reflection depends on localizing the failure before revision, rather than having the verifier assess the entire inference trajectory at once.

\paragraph{Answer Transitions after Recovery.}
We compare the initial supplementary answer $\hat{y}_0$ with the final answer only for instances that trigger at least one recovery action. For this transition analysis, numerical answers are considered
correct if their relative error with respect to the ground truth
is at most $25\%$. Recovery is invoked for roughly 30\% of each model's refinement subset. Among these instances, about two-thirds of transitions are wrong-to-correct, whereas correct-to-wrong transitions remain below 3\% for all three models (Fig.~\ref{fig:reflection-outcomes}). These transitions indicate that revisions usually repair an incorrect initial answer while rarely overturning a correct answer.


\par\endgroup

\subsection{Recovery Cost and Budget}

\paragraph{Selective Recovery versus Full Restart.}
To isolate recovery scope, both strategies start from the same state and use identical stage-aligned verification with $T=2$. Full restart reruns the entire refinement trajectory from the fixed native state, including planning, perception, and reasoning. Selective recovery recomputes only invalidated states and their dependencies. Selective recovery reduces mean model calls for all three models (Fig.~\ref{fig:correction-efficiency}; counting rule in Appendix~\ref{app:exp-protocol}). Qwen and GPT improve in both cost and score, whereas Gemini exhibits a small cost--accuracy trade-off.

\paragraph{Recovery Budget.}
As shown in Fig.~\ref{fig:correction-efficiency}, increasing $T$ from 1 to 2 improves the overall score for every base model. Raising the budget to 3 increases model calls for all three, but yields only a small gain for Qwen and lower scores for Gemini and GPT. Thus, a larger recovery budget does not consistently improve performance.

\subsection{Qualitative Analysis}
\label{sec:qualitative-results}

\begin{figure}[t]
\centering
\setlength{\abovecaptionskip}{4pt}
\includegraphics[width=\linewidth]{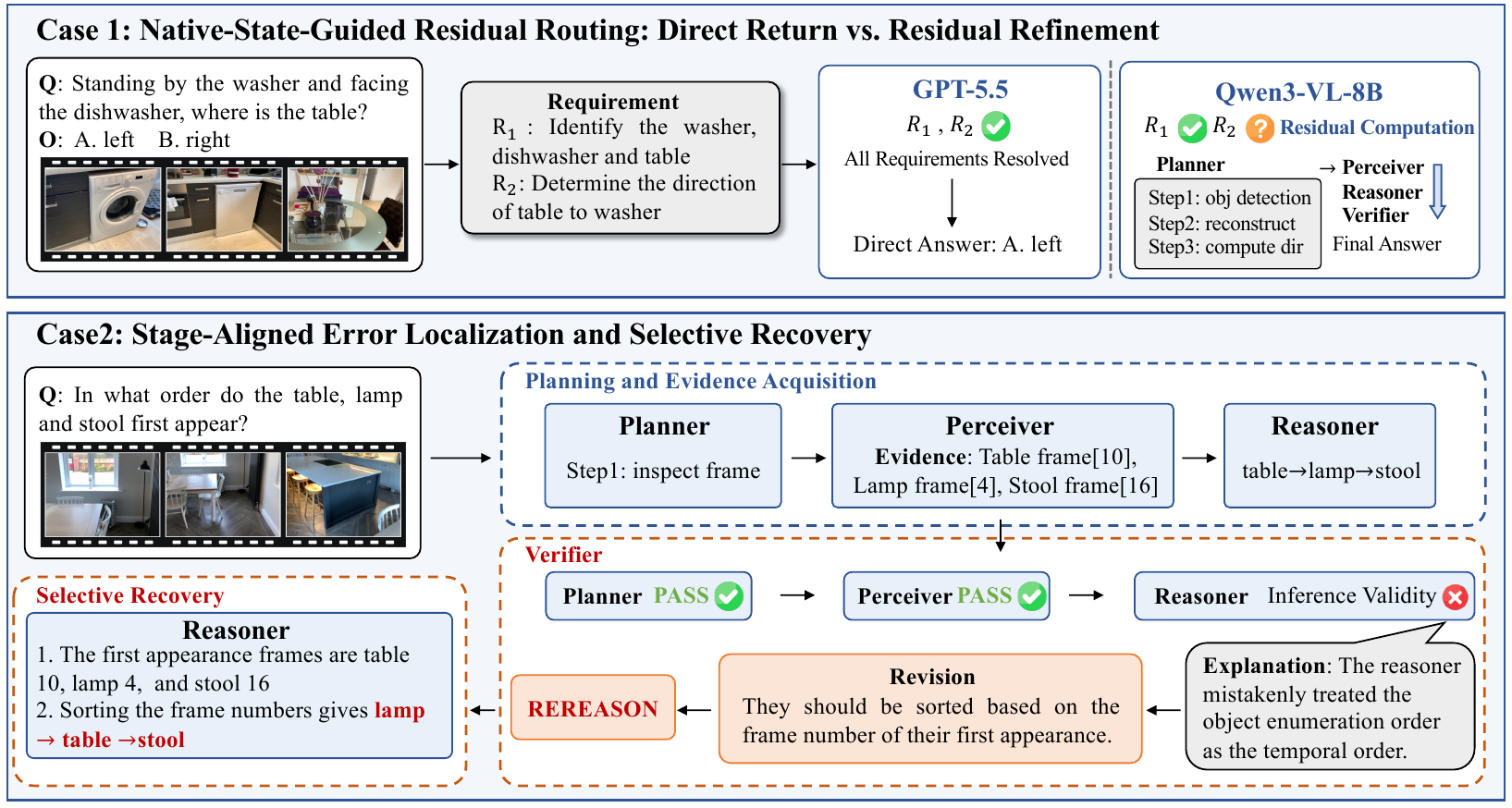}
\vspace{-0.4cm}
\caption{\textbf{Qualitative examples of residual allocation and recovery.}
\textbf{Top:} The same query follows different routes because GPT-5.5 resolves both native requirements, whereas Qwen3-VL-8B leaves the directional relation unresolved.
\textbf{Bottom:} Verification localizes an appearance-order error to reasoning. \texttt{REREASON} corrects the order while reusing the verified plan and evidence.}
\label{fig:case-study}
\vspace{-2pt}
\end{figure}

Fig.~\ref{fig:case-study} shows the two residual decisions: compute only for unresolved requirements, then recompute only invalidated states and their dependents.

\paragraph{Case 1: Native-State-Guided Residual Routing.}
The query requires identifying the washer, dishwasher, and table ($R_1$) and resolving the table's direction in the specified observer frame ($R_2$). GPT-5.5 supports both requirements and returns its native candidate, \emph{A: left}. Qwen3-VL-8B supports $R_1$ but leaves $R_2$ unresolved, so refinement adds detection, reconstruction, and direction computation. The supported identity evidence remains active rather than being reacquired. The same query follows different execution paths across base models, illustrating that residual computation is determined by the current model--instance state rather than by the task alone.

\paragraph{Case 2: Stage-Aligned Error Localization and Selective Recovery.}
For the appearance-order query, the Perceiver records first appearances at frames 10, 4, and 16 for the table, lamp, and stool, but the Reasoner preserves the enumeration order and returns \emph{table $\rightarrow$ lamp $\rightarrow$ stool}. Planning and perception pass verification, while the inference-validity check  localizes the error to reasoning. Guided by the Verifier's revision to sort objects by their first-appearance frames, ReTAC applies \texttt{REREASON} without changing $P_t$ or $\mathcal{E}_t$.  The retained frame indices, $4<10<16$, establish the corrected order \emph{lamp $\rightarrow$ table $\rightarrow$ stool} without new perception. This correction updates only the reasoning trace and answer, illustrating dependency-based state reuse.

\section{Conclusion}
\label{sec:conclusion}

We introduced ReTAC, a training-free multiagent framework that formulates test-time agentic computation for video spatial reasoning as residual computation: supplementary computation targets only requirements unresolved by native inference, while recovery recomputes only invalidated states and their dependencies. Across three MLLMs and two benchmarks, ReTAC achieves the highest overall score among the compared methods. Controlled routing interventions show that forcing refinement on direct-return instances or suppressing refinement on refinement instances reduces performance.  Selective recovery reduces mean model calls relative to full restart with comparable overall performance. The results support model--instance specific computation allocation and dependency-based state reuse over repeatedly executing a full agentic pipeline.

\section{Future Work}
\label{future work}
ReTAC's practical scope is set primarily by the quality of its state estimates and supplementary operations. A native call that incorrectly marks incomplete evidence as supported may terminate early. Future work will develop calibrated support estimates and uncertainty-aware routing criteria. Additionally, the supplementary evidence is likewise bounded by the operation interface. For example, VGGT-based geometry does not provide metric
scale, which is consistent with the less uniform gains on metric tasks. Future work will extend the interface with more robust spatial
operations.

\newpage
\bibliography{iclr2027_conference}

@article{mou2026visra,
  title={ViSRA: A Video-based Spatial Reasoning Agent for Multi-modal Large Language Models},
  author={Mou, Tingshu and He, Jiabo and Wang, Renying and Liu, Ce and Yang, Hao and Zhang, Tiehua and Chen, Jingjing and Ma, Xingjun},
  journal={arXiv preprint arXiv:2605.10106},
  year={2026}
}

@inproceedings{
li2025seetrek,
title={See\&Trek: Training-Free Spatial Prompting for Multimodal Large Language Model},
author={Pengteng Li and Pinhao Song and Wuyang Li and Huizai Yao and Weiyu Guo and Yijie Xu and Dugang Liu and Hui Xiong},
booktitle={The Thirty-ninth Annual Conference on Neural Information Processing Systems},
year={2025},
url={https://openreview.net/forum?id=2exr4mlbx1}
}

@inproceedings{hua2026unleashing,
  title={Unleashing spatial reasoning in multimodal large language models via textual representation guided reasoning},
  author={Hua, Jiacheng and Yin, Yishu and Wu, Yuhang and Wang, Tai and Huang, Yifei and Liu, Miao},
  booktitle={Proceedings of the 64th Annual Meeting of the Association for Computational Linguistics (Volume 1: Long Papers)},
  pages={13616--13637},
  year={2026}
}

@article{zhang2026think3d,
  title={Think3d: Thinking with space for spatial reasoning},
  author={Zhang, Zaibin and Wu, Yuhan and Jia, Lianjie and Wang, Yifan and Zhang, Zhongbo and Li, Yijiang and Ran, Binghao and Zhang, Fuxi and Sun, Zhuohan and Yin, Zhenfei and others},
  journal={arXiv preprint arXiv:2601.13029},
  year={2026}
}

@article{ruan2026allospatial,
  title={AlloSpatial: Agentic Harness Framework for Spatial Reasoning in Foundation Models},
  author={Ruan, Shouwei and Wang, Bin and Wu, Zhenyu and Zhu, Qihui and Zhang, Yuxiang and Li, Jingzhi and Wang, Yubin and Wei, Xingxing},
  journal={arXiv preprint arXiv:2606.08952},
  year={2026}
}

@inproceedings{fan2026vlm,
  title={Vlm-3r: Vision-language models augmented with instruction-aligned 3d reconstruction},
  author={Fan, Zhiwen and Zhang, Jian and Li, Renjie and Zhang, Junge and Chen, Runjin and Hu, Hezhen and Wang, Kevin and Wang, Peihao and Qu, Huaizhi and Zhou, Shijie and others},
  booktitle={Proceedings of the IEEE/CVF Conference on Computer Vision and Pattern Recognition},
  pages={31054--31065},
  year={2026}
}

@article{cheng2026grounded,
  title={Grounded 3D-Aware Spatial Vision-Language Modeling},
  author={Cheng, An-Chieh and Fu, Yang and Ji, Yatai and Zhu, Ligeng and Zhan, Guanqi and Zhang, Zhuoyang and Yang, Zhaojing and Han, Song and Lu, Yao and Molchanov, Pavlo and others},
  journal={arXiv preprint arXiv:2605.30307},
  year={2026}
}

@inproceedings{qi2026gpt4scene,
  title={Gpt4scene: Understand 3d scenes from videos with vision-language models},
  author={Qi, Zhangyang and Zhang, Zhixiong and Fang, Ye and Wang, Jiaqi and Zhao, Hengshuang},
  booktitle={International Conference on Learning Representations},
  volume={2026},
  pages={157277--157302},
  year={2026}
}

@inproceedings{wang2026adatooler,
  title={Adatooler-v: Adaptive tool-use for images and videos},
  author={Wang, Chaoyang and Feng, Kaituo and Chen, Dongyang and Wang, Zhongyu and Li, Zhixun and Gao, Sicheng and Meng, Meng and Zhou, Xu and Zhang, Manyuan and Shang, Yuzhang and others},
  booktitle={Findings of the Association for Computational Linguistics: ACL 2026},
  pages={18058--18072},
  year={2026}
}

@article{li2026skill,
  title={Skill-3D: Evolving Scene-Aware Skills for Agentic 3D Spatial Reasoning},
  author={Li, Haoyuan and Hu, Zhengdong and Wang, Jun and Fan, Hehe and Yang, Yi},
  journal={arXiv preprint arXiv:2606.07436},
  year={2026}
}

@article{dai2026s,
  title={S-Agent: Spatial Tool-Use Elicits Reasoning for Spatial Intelligence},
  author={Dai, Yalun and Li, Hao and Tian, Shulin and Yao, Runmao and Dong, Yuhao and Hong, Fangzhou and Chen, Zhaoxi and Liu, Fangfu and Tian, Baoliang and Zhang, Dingwen and others},
  journal={arXiv preprint arXiv:2606.20515},
  year={2026}
}

@InProceedings{Dongfang_2026_CVPR,
    author    = {Dongfang, Zihao and Zheng, Xu and Weng, Ziqiao and Lyu, Yuanhuiyi and Paudel, Danda Pani and Van Gool, Luc and Yang, Kailun and Hu, Xuming},
    title     = {Are Multimodal Large Language Models Ready for Omnidirectional Spatial Reasoning?},
    booktitle = {Proceedings of the IEEE/CVF Conference on Computer Vision and Pattern Recognition (CVPR) Findings},
    month     = {June},
    year      = {2026},
    pages     = {9759-9769}
}

@article{li2026ovo,
  title={OVO-S-Bench: A Hierarchical Benchmark for Streaming Spatial Intelligence in Multimodal LLMs},
  author={Li, Yifei and Liu, Pengyiang and Zang, Yuhang and Shi, Zhongyue and Fu, Qi and Hao, Hongye and Lu, Jiwen},
  journal={arXiv preprint arXiv:2606.03890},
  year={2026}
}

@article{yang2025embodiedbench,
  title={Embodiedbench: Comprehensive benchmarking multi-modal large language models for vision-driven embodied agents},
  author={Yang, Rui and Chen, Hanyang and Zhang, Junyu and Zhao, Mark and Qian, Cheng and Wang, Kangrui and Wang, Qineng and Koripella, Teja Venkat and Movahedi, Marziyeh and Li, Manling and others},
  journal={arXiv preprint arXiv:2502.09560},
  year={2025}
}

@inproceedings{yang2025thinking,
  title={Thinking in space: How multimodal large language models see, remember, and recall spaces},
  author={Yang, Jihan and Yang, Shusheng and Gupta, Anjali W and Han, Rilyn and Fei-Fei, Li and Xie, Saining},
  booktitle={2025 IEEE/CVF Conference on Computer Vision and Pattern Recognition (CVPR)},
  pages={10632--10643},
  year={2025},
  organization={IEEE}
}

@inproceedings{li2025sti,
  title={Sti-bench: Are mllms ready for precise spatial-temporal world understanding?},
  author={Li, Yun and Zhang, Yiming and Lin, Tao and Liu, XiangRui and Cai, Wenxiao and Liu, Zheng and Zhao, Bo},
  booktitle={2025 IEEE/CVF International Conference on Computer Vision (ICCV)},
  pages={5622--5632},
  year={2025},
  organization={IEEE}
}

@inproceedings{liu2024grounding,
  title={Grounding dino: Marrying dino with grounded pre-training for open-set object detection},
  author={Liu, Shilong and Zeng, Zhaoyang and Ren, Tianhe and Li, Feng and Zhang, Hao and Yang, Jie and Jiang, Qing and Li, Chunyuan and Yang, Jianwei and Su, Hang and others},
  booktitle={European conference on computer vision},
  pages={38--55},
  year={2024},
  organization={Springer}
}

@inproceedings{wang2025vggt,
  title={Vggt: Visual geometry grounded transformer},
  author={Wang, Jianyuan and Chen, Minghao and Karaev, Nikita and Vedaldi, Andrea and Rupprecht, Christian and Novotny, David},
  booktitle={2025 IEEE/CVF Conference on Computer Vision and Pattern Recognition (CVPR)},
  pages={5294--5306},
  year={2025},
  organization={IEEE}
}

@inproceedings{song2026adareasoner,
  title={Adareasoner: Dynamic tool orchestration for iterative visual reasoning},
  author={Song, Mingyang and Sun, Haoyu and Gu, Jiawei and Li, Linjie and Krishna, Ranjay and Cheng, Yu},
  booktitle={International Conference on Learning Representations},
  volume={2026},
  pages={142731--142754},
  year={2026}
}

@inproceedings{sivakumaran2026dart,
  title={Dart: Leveraging multi-agent disagreement for tool recruitment in multimodal reasoning},
  author={Sivakumaran, Nithin and Chen, Justin and Wan, David and Zhang, Yue and Yoon, Jaehong and Stengel-Eskin, Elias and Bansal, Mohit},
  booktitle={Proceedings of the 19th Conference of the European Chapter of the Association for Computational Linguistics (Volume 1: Long Papers)},
  pages={5445--5464},
  year={2026}
}

@inproceedings{deng2026toolscope,
  title={ToolScope: An Agentic Framework for Vision-Guided and Long-Horizon Tool Use},
  author={Deng, Mengjie and Dong, Guanting and Dou, Zhicheng},
  booktitle={Findings of the Association for Computational Linguistics: ACL 2026},
  pages={211--225},
  year={2026}
}

@article{jiang2026refer,
  title={Refer-agent: A collaborative multi-agent system with reasoning and reflection for referring video object segmentation},
  author={Jiang, Haichao and Liang, Tianming and Zheng, Wei-Shi and Hu, Jian-Fang},
  journal={arXiv preprint arXiv:2602.03595},
  year={2026}
}

@inproceedings{zhan2026gthinker,
  title={Gthinker: Towards general multimodal reasoning via cue-guided rethinking},
  author={Zhan, Yufei and Wu, Ziheng and Zhu, Yousong and Xue, Rongkun and Zhou, Guanghao and Luo, Ruipu and Chen, Zhenghao and Zhang, Can and Li, Yifan and He, Zhentao and others},
  booktitle={Proceedings of the IEEE/CVF Conference on Computer Vision and Pattern Recognition},
  pages={11954--11965},
  year={2026}
}

@inproceedings{li2026revisor,
  title={Revisor: Beyond textual reflection, towards multimodal introspective reasoning in long-form video understanding},
  author={Li, Jiaze and Yin, Hao and Tan, Wenhui and Chen, Jingyang and Xu, Boshen and Qu, Yuxun and Chen, Yijing and Ju, Jianzhong and Luo, Zhenbo and Luan, Jian},
  booktitle={Proceedings of the IEEE/CVF Conference on Computer Vision and Pattern Recognition},
  pages={5059--5069},
  year={2026}
}

@inproceedings{wu2025visco,
  title={Visco: Benchmarking fine-grained critique and correction towards self-improvement in visual reasoning},
  author={Wu, Xueqing and Ding, Yuheng and Li, Bingxuan and Lu, Pan and Yin, Da and Chang, Kai-Wei and Peng, Nanyun},
  booktitle={2025 IEEE/CVF Conference on Computer Vision and Pattern Recognition (CVPR)},
  pages={9527--9537},
  year={2025},
  organization={IEEE}
}

@misc{brown2025benchmarkdesignerstraintest,
      title={Benchmark Designers Should "Train on the Test Set" to Expose Exploitable Non-Visual Shortcuts}, 
      author={Ellis Brown and Jihan Yang and Shusheng Yang and Rob Fergus and Saining Xie},
      year={2025},
      eprint={2511.04655},
      archivePrefix={arXiv},
      primaryClass={cs.CV},
      url={https://arxiv.org/abs/2511.04655}, 
}

@inproceedings{snell2025scaling,
  title={Scaling LLM test-time compute optimally can be more effective than scaling parameters for reasoning},
  author={Snell, Charlie and Lee, Jaehoon and Xu, Kelvin and Kumar, Aviral},
  booktitle={International Conference on Learning Representations},
  volume={2025},
  pages={10131--10165},
  year={2025}
}

@article{bai2025qwen3,
  title={Qwen3-vl technical report},
  author={Bai, Shuai and Cai, Yuxuan and Chen, Ruizhe and Chen, Keqin and Chen, Xionghui and Cheng, Zesen and Deng, Lianghao and Ding, Wei and Gao, Chang and Ge, Chunjiang and others},
  journal={arXiv preprint arXiv:2511.21631},
  year={2025}
}

@inproceedings{dai2017scannet,
  title={Scannet: Richly-annotated 3d reconstructions of indoor scenes},
  author={Dai, Angela and Chang, Angel X and Savva, Manolis and Halber, Maciej and Funkhouser, Thomas and Nie{\ss}ner, Matthias},
  booktitle={2017 IEEE conference on computer vision and pattern recognition (CVPR)},
  pages={2432--2443},
  year={2017},
  organization={IEEE}
}

@inproceedings{yeshwanth2023scannet++,
  title={Scannet++: A high-fidelity dataset of 3d indoor scenes},
  author={Yeshwanth, Chandan and Liu, Yueh-Cheng and Nie{\ss}ner, Matthias and Dai, Angela},
  booktitle={2023 IEEE/CVF International Conference on Computer Vision (ICCV)},
  pages={12--22},
  year={2023},
  organization={IEEE}
}

@article{baruch2021arkitscenes,
  title={Arkitscenes: A diverse real-world dataset for 3d indoor scene understanding using mobile rgb-d data},
  author={Baruch, Gilad and Chen, Zhuoyuan and Dehghan, Afshin and Dimry, Tal and Feigin, Yuri and Fu, Peter and Gebauer, Thomas and Joffe, Brandon and Kurz, Daniel and Schwartz, Arik and others},
  journal={arXiv preprint arXiv:2111.08897},
  year={2021}
}

@inproceedings{zhang2024omni6dpose,
  title={Omni6dpose: A benchmark and model for universal 6d object pose estimation and tracking},
  author={Zhang, Jiyao and Huang, Weiyao and Peng, Bo and Wu, Mingdong and Hu, Fei and Chen, Zijian and Zhao, Bo and Dong, Hao},
  booktitle={European Conference on Computer Vision},
  pages={199--216},
  year={2024},
  organization={Springer}
}

@inproceedings{sun2020scalability,
  title={Scalability in perception for autonomous driving: Waymo open dataset},
  author={Sun, Pei and Kretzschmar, Henrik and Dotiwalla, Xerxes and Chouard, Aurelien and Patnaik, Vijaysai and Tsui, Paul and Guo, James and Zhou, Yin and Chai, Yuning and Caine, Benjamin and others},
  booktitle={2020 IEEE/CVF Conference on Computer Vision and Pattern Recognition (CVPR)},
  pages={2443--2451},
  year={2020},
  organization={IEEE}
}

@misc{sun2026restisourcegroundedauditrepair,
      title={ReSTI: A Source-Grounded Audit and Repair of STI-Bench}, 
      author={Pengzhan Sun and Ramanathan Rajaraman and Shiu-Hong Kao and Junbin Xiao and Angela Yao},
      year={2026},
      eprint={2609.24727},
      archivePrefix={arXiv},
      primaryClass={cs.CV},
      url={https://arxiv.org/abs/2609.24727}, 
}

@inproceedings{ashraf2026agent,
  title={Agent-x: Evaluating deep multimodal reasoning in vision-centric agentic tasks},
  author={Ashraf, Tajamul and Saqib, Amal and Gani, Hanan and AlMahri, Muhra and Li, Yuhao and Ahsan, Noor and Nawaz, Umair and Lahoud, Jean and Cholakkal, Hisham and Shah, Mubarak and others},
  booktitle={International Conference on Learning Representations},
  volume={2026},
  pages={49814--49874},
  year={2026}
}

@inproceedings{NEURIPS2025_2c84844a,
 author = {Wang, Haozhe and Qu, Chao and Huang, Zuming and Chu, Wei and Lin, Fangzhen and Chen, Wenhu},
 booktitle = {Advances in Neural Information Processing Systems},
 doi = {10.52202/085713-1037},
 editor = {D. Belgrave and C. Zhang and H. Lin and R. Pascanu and P. Koniusz and M. Ghassemi and N. Chen},
 pages = {30865--30891},
 publisher = {Curran Associates, Inc.},
 title = {VL-Rethinker: Incentivizing Self-Reflection of Vision-Language Models with Reinforcement Learning},
 url = {https://proceedings.neurips.cc/paper_files/paper/2025/file/2c84844a559e4f962752570bff456ae4-Paper-Conference.pdf},
 volume = {38, Main Conference},
 year = {2025}
}

@article{wang2025visualprm,
  title={Visualprm: An effective process reward model for multimodal reasoning},
  author={Wang, Weiyun and Gao, Zhangwei and Chen, Lianjie and Chen, Zhe and Zhu, Jinguo and Zhao, Xiangyu and Liu, Yangzhou and Cao, Yue and Ye, Shenglong and Zhu, Xizhou and others},
  journal={arXiv preprint arXiv:2503.10291},
  year={2025}
}
\bibliographystyle{iclr2027_conference}

\newpage

\appendix

\begin{center}
  {\LARGE\scshape Supplementary Material\par}
  \vspace{0.6em}
  {\Large\bfseries ReTAC: Residual Test-Time Agentic Computation for Video Spatial Reasoning\par}
\end{center}

\section{Residual Inference, Stage-Aligned Verification, and Selective Recovery}
\label{app:method-details}

This appendix specifies the state semantics, stage-specific verification inputs, and recovery rules underlying Sections~\ref{sec:overview}--\ref{sec:rollback}. We retain the notation of the main text and focus on the details needed to reproduce residual routing, verification, and dependency-aware state reuse.

\subsection{State Representation and Evidence Integration}
\label{app:state-details}

\paragraph{Native State.}
The native call in Eq.~\ref{eq:overview-native} receives only $x=(V,q,\mathcal{O})$, without the operation interface or tool outputs. Each $r\in\mathcal{R}^{\mathrm{nat}}$ is labeled \texttt{SUPPORTED} or \texttt{UNRESOLVED}, and the unresolved set is
\begin{equation}
\mathcal{U}^{\mathrm{nat}}
=
\{r\in\mathcal{R}^{\mathrm{nat}}:
\operatorname{status}(r)=\texttt{UNRESOLVED}\}.
\label{eq:native-gaps}
\end{equation}
These labels determine routing; they are not verification outcomes. Because the Planner is invoked only after an instance is routed to refinement, requirements omitted by the native call can be introduced only on that branch.

\paragraph{Residual Plan.}
For $\mathcal{U}^{\mathrm{nat}}\neq\varnothing$, the Planner conditions on the native state and the documented interface:
\begin{equation}
P_0=(\mathcal{S}_0,\mathcal{R}_0,\mathcal{U}_0)
=\operatorname{Plan}_M(x,S^{\mathrm{nat}},\mathcal{T}).
\label{eq:residual-plan}
\end{equation}

It may add requirements omitted by native inference, with
$\mathcal{R}_0$ and $\mathcal{U}_0$ defined by Eq.~\ref{eq:residual-requirements}. Each step $s_k \in \mathcal{S}_0 = (s_1,\ldots,s_K)$ specifies
a target requirement, a primitive operation or a composition
of primitive operations, arguments, input references,
dependencies, and expected output.
Operations follow dependency order both within and across steps.
Steps are included only when they resolve an element of
$\mathcal{U}_0$ or provide a prerequisite for doing so.
Thus, $\mathcal{U}_0$ records acquisition targets at planning time
rather than verified resolutions.
The number of steps $K$ is determined by the Planner and has
no fixed upper bound.

\paragraph{Evidence Records and Integration.}
Native and supplementary observations share the record structure in Eq.~\ref{eq:evidence}. The source field $\sigma_j$ identifies the frame indices, raw operation record, or record field supporting $o_j$, whereas $\mathcal{D}_j$ records the inputs used to produce the observation. For an observation produced by $s_k$, $r_j=\operatorname{req}(s_k)$ and $p_j=s_k$; native observations use $p_j=\mathrm{nat}$.

The Perceiver resolves the declared inputs and executes each step without altering its operation, arguments, or acquisition scope. Failed acquisitions are logged without creating an observation. If $\mathcal{E}^{\mathrm{new}}_k$ denotes the records produced by $s_k$, then
\begin{equation}
\mathcal{E}^{\mathrm{new}}
=\operatorname{Perceive}_M(x,P_0,\mathcal{T})
=\bigcup_{k=1}^{K}\mathcal{E}^{\mathrm{new}}_k,
\label{eq:supp-evidence}
\end{equation}
and the initial active evidence set is
\begin{equation}
\mathcal{E}_0
=\operatorname{Merge}(\mathcal{E}^{\mathrm{nat}},\mathcal{E}^{\mathrm{new}}).
\label{eq:evidence-merge}
\end{equation}
The merge operation does not collapse conflicting observations: each record remains associated with its source and dependencies. Raw outputs are retained in the execution history even when a derived observation is superseded. During recovery, the corrected record replaces its predecessor only in the active evidence set.

\paragraph{Evidence-Grounded Reasoning.}
The Reasoner uses $(q, O, P_0, \mathcal{E}_0)$ to produce
$(Z_0, \hat{y}_0)$. Every critical claim in $Z_0$ must reference active
evidence or a preceding supported claim. The Reasoner cannot introduce
new visual observations. Any additional visual observation must be
acquired by the Perceiver and registered with its source in the active
evidence set before the Reasoner can use it. The plan specifies what
information must be resolved but does not itself constitute evidence.

\subsection{Stage-Aligned Verification Details}
\label{app:verification-details}

For $\Sigma_t=(P_t,\mathcal{E}_t,Z_t,\hat{y}_t)$, verification proceeds in the order planning $\rightarrow$ perception $\rightarrow$ reasoning. The model evaluates semantic support, while deterministic checks validate interfaces, references, and traceable computations.

\paragraph{Outcomes and Aggregation.}
Each criterion returns \texttt{PASS}, \texttt{FAIL}, or \texttt{UNKNOWN} and includes a target and justification. Equation~\ref{eq:verification-aggregation} gives the stage aggregation rule: a stage passes only if every criterion passes, and \texttt{UNKNOWN} is non-passing. Verification stops at the first non-passing stage, so later stages are unevaluated rather than assigned \texttt{UNKNOWN}.

\paragraph{Planning.}
\label{sec:verify-planning}
The planning check receives only $(q,\mathcal{O},P_t,\mathcal{T})$; it does not access the video, observation contents, reasoning trace, or candidate answer. Deterministic checks use declared metadata for references and dependencies.

\emph{Plan completeness.} The plan must cover the entities, relations, and spatial or temporal conditions needed to answer the question. A missing acquisition is necessary when the information it would provide could change the answer while the represented requirements are held fixed. An identifiable omission yields \texttt{FAIL}; insufficient information to determine completeness yields \texttt{UNKNOWN}.

\emph{Operation feasibility.} Deterministic checks validate registered primitives, required
arguments, types and ranges, input references, and dependencies
both within composite steps and across plan steps. The semantic check assesses whether the documented outputs and limitations of the selected operations can support their target requirements. Invalid interfaces or incompatible acquisition procedures yield \texttt{FAIL}; indeterminate feasibility yields \texttt{UNKNOWN}. A pass establishes feasibility, not correctness of the acquired observation.

\paragraph{Perception.}
\label{sec:verify-perception}
This stage receives $P_t$, $\mathcal{E}_t$, the cited frames, raw operation outputs, and the relevant interface definitions. Native and supplementary records are evaluated by the same criteria.

\emph{Source validity.} Reference checks validate frame indices, operation records, and cited output fields. The semantic check then determines whether the cited source supports the observation under the specified object identity and spatial or temporal context. A definite inconsistency yields \texttt{FAIL}; inaccessible or ambiguous support yields \texttt{UNKNOWN}.

\emph{Requirement coverage.} Every $r\in\mathcal{R}_t$ must be resolved by valid active evidence. Missing or contradicted evidence yields \texttt{FAIL}; unresolved support or relevance yields \texttt{UNKNOWN}. Neither the number of acquired records nor the Perceiver's execution status determines coverage.

\paragraph{Reasoning.}
\label{sec:verify-reasoning}
This stage receives $(q,\mathcal{O},P_t,\mathcal{E}_t,Z_t,\hat{y}_t)$ and checks the derivation from active evidence to the returned answer.

\emph{Evidence support.} Reference checks require every critical claim to point to active evidence or preceding supported claims; semantic checks verify that those sources support the stated premises. Invalid references, contradicted premises, or identifiable unsupported assumptions yield \texttt{FAIL}; ambiguous support yields \texttt{UNKNOWN}.

\emph{Inference validity.} Deterministic checks are used when the operators and operands are traceable to active evidence; other semantic or geometric inferences are evaluated by the model. A definite logical or computational error yields \texttt{FAIL}; an inference that cannot be verified or rejected yields \texttt{UNKNOWN}.

\emph{Answer consistency.} The returned answer must agree with the conclusion in $Z_t$. For multiple-choice questions, labels or exact option text are mapped to option identifiers. Numerical answers are normalized under the required unit conversion and rounding rules. Disagreement, an invalid output format, or an incorrect mapping yields \texttt{FAIL}; an ambiguous mapping yields \texttt{UNKNOWN}.

\subsection{Selective Recovery and Cache Invalidation}
\label{app:recovery-details}

\paragraph{Recovery Routing and Upstream Rechecks.}
Let $J_t$ denote the verifier's proposed correction, containing a target, diagnosis, and revision. Given the stage outcomes $\boldsymbol{v}_t$, the controller produces
\begin{equation}
F_t=(a_t,\tau_t,\delta_t,\rho_t)
=\operatorname{Route}(\boldsymbol{v}_t,J_t),
\label{eq:recovery-route}
\end{equation}
where $a_t$ is the action, $\tau_t$ the target, $\delta_t$ the diagnosis, and $\rho_t$ the proposed revision. \textsc{Route} returns \texttt{ACCEPT} only when all three stages pass. Otherwise, it selects a recovery operator only when the diagnosis specifies a concrete revision; if not, it returns $\varnothing$.

A later-stage diagnosis may raise a specific concern about an earlier pass. Before recovery is selected, only the implicated upstream criterion is rechecked. A confirmed violation changes that criterion to \texttt{FAIL}, unresolved support to \texttt{UNKNOWN}, and a refuted concern leaves the cached \texttt{PASS} unchanged.

For a composite step, changing its internal operations,
arguments, or dependencies requires REPLAN.
REPERCEIVE repeats the existing composition without
changing its acquisition definition.

\paragraph{Recovery Operators.}
\texttt{REPLAN} revises requirements or acquisition steps, including their operation, arguments, inputs, or dependencies. Evidence is retained only when its object identity, temporal and spatial references, scale, and acquisition conditions remain compatible with the revised plan. \texttt{REPERCEIVE} keeps the plan fixed and repeats the affected acquisition with the same operation, arguments, inputs, and dependencies; any change to that acquisition definition is handled by \texttt{REPLAN}. \texttt{REREASON} preserves both the plan and active evidence and regenerates the derivation and answer.

\paragraph{Dependency-Based Invalidation.}
Let $\Delta_t$ contain the records or steps directly changed or invalidated by recovery. The affected set is
\begin{equation}
\mathcal{I}_t
=\Delta_t\cup\operatorname{Desc}(\Delta_t),
\end{equation}
where $\operatorname{Desc}$ is the transitive closure under recorded input and support references. Superseded observations are removed from the active set but retained in the execution history. A cached criterion can be reused only if its supporting records are outside $\mathcal{I}_t$ and its justification remains valid for the revised state. Because every recovery generates a new reasoning trace and answer $(Z_{t+1},\hat{y}_{t+1})$, checks on the previous reasoning or answer are never reused. The history $H$ stores recovery feedback, whereas $C$ stores criterion outcomes and their dependencies.

\paragraph{Budget Semantics.}
The budget $T$ counts recovery actions rather than verification rounds. The initial state is verified at $t=0$, and the state produced by the final permitted action is verified at $t=T$. A state that passes at $t=T$ is therefore returned as \texttt{VERIFIED}. If verification does not support another revision, or if the final state remains non-passing, ReTAC returns the current answer as \texttt{UNVERIFIED}. \texttt{VERIFIED} denotes acceptance under these criteria rather than a ground-truth label.

\subsection{Complete Inference Procedure}
\label{app:retac-procedure}

Algorithm~\ref{alg:retac} combines native routing, residual refinement, ordered verification, and bounded recovery. \textsc{Verify} applies the stage-specific inputs above and reuses only valid cache entries; \textsc{ResolveConcerns} performs targeted upstream rechecks when needed. \textsc{Recover} applies $F_t$ and returns the revised plan, active evidence, and directly invalidated items $\Delta_t$. During a recovery iteration, the Reasoner additionally receives the accumulated history $H$ and the current feedback $F_t$; we denote this conditioning after a semicolon.

\begin{algorithm}[t]
\caption{\textbf{ReTAC inference with bounded selective recovery.}}
\label{alg:retac}
\small
\begin{algorithmic}[1]
\Require $x=(V,q,\mathcal{O})$, model $M$, interface $\mathcal{T}$, recovery budget $T$
\Ensure answer $\hat{y}$ and status $s$
\State $S^{\mathrm{nat}}\gets\operatorname{Native}_M(x)$
\If{$\mathcal{U}^{\mathrm{nat}}=\varnothing$}
    \State \Return $(\hat{y}^{\mathrm{nat}},\texttt{DIRECT})$
\EndIf
\State $P_0\gets\operatorname{Plan}_M(x,S^{\mathrm{nat}},\mathcal{T})$
\State $\mathcal{E}^{\mathrm{new}}\gets\operatorname{Perceive}_M(x,P_0,\mathcal{T})$
\State $\mathcal{E}_0\gets\operatorname{Merge}(\mathcal{E}^{\mathrm{nat}},\mathcal{E}^{\mathrm{new}})$
\State $(Z_0,\hat{y}_0)\gets\operatorname{Reason}_M(q,\mathcal{O},P_0,\mathcal{E}_0)$
\State $H\gets()$, $C\gets\varnothing$
\For{$t=0,\ldots,T$}
    \State $\Sigma_t\gets(P_t,\mathcal{E}_t,Z_t,\hat{y}_t)$
    \State $(\boldsymbol{v}_t,J_t,C)\gets\Call{Verify}{x,\mathcal{T},\Sigma_t,C}$
    \State $(\boldsymbol{v}_t,J_t,C)\gets\Call{ResolveConcerns}{x,\mathcal{T},\Sigma_t,\boldsymbol{v}_t,J_t,C}$
    \State $F_t=(a_t,\tau_t,\delta_t,\rho_t)\gets\operatorname{Route}(\boldsymbol{v}_t,J_t)$
    \If{$a_t=\texttt{ACCEPT}$}
        \State \Return $(\hat{y}_t,\texttt{VERIFIED})$
    \EndIf
    \If{$a_t=\varnothing$ \textbf{or} $t=T$}
        \State \Return $(\hat{y}_t,\texttt{UNVERIFIED})$
    \EndIf
    \State $H\gets\operatorname{Append}(H,F_t)$
    \State $(P_{t+1},\mathcal{E}_{t+1},\Delta_t)\gets
       \Call{Recover}{x,M,\mathcal{T},P_t,\mathcal{E}_t,H,F_t}$
    \State $\Delta_t\gets\Delta_t\cup\{Z_t,\hat{y}_t\}$
    \State $\mathcal{I}_t\gets\Delta_t\cup\operatorname{Desc}(\Delta_t)$
    \State $C\gets\operatorname{Invalidate}(C,\mathcal{I}_t)$
    \State $(Z_{t+1},\hat{y}_{t+1})\gets
       \operatorname{Reason}_M(q,\mathcal{O},P_{t+1},\mathcal{E}_{t+1};H,F_t)$
\EndFor
\end{algorithmic}
\end{algorithm}

\section{Evaluation and Controlled Experimental Protocols}
\label{app:exp-details}

\subsection{Shared Evaluation Protocol}
\label{app:exp-protocol}

We evaluate Qwen3-VL-8B, Gemini~3.1~Pro, and GPT-5.5 on VSI-Bench-Debiased and STI-Bench, using 32 uniformly sampled frames per video. The native call receives only the sampled video, question, and optional answer choices, without tool descriptions or outputs. All tool-enabled variants share the operation interface in Appendix~\ref{app:tool_definitions}; the same base MLLM instantiates the Planner, Perceiver, Reasoner, and Verifier under role-specific instructions fixed across models (Appendix~\ref{app:agent-prompts}).

The recovery budget $T$ counts recovery actions rather than verification rounds: the initial state is verified at $t=0$, and every recovered state is verified through the final permitted action at $t=T$. Unless stated otherwise, $T=2$. All routing, component-replacement, answer-transition, and recovery-budget analyses use VSI-Bench-Debiased.

For the recovery-efficiency analyses on VSI-Bench-Debiased, we report
the number of base-MLLM invocations averaged over the full
2,362-question evaluation split, including instances returned directly
after native inference. For instance $i$, $C_i$ counts the initial
native call; all Planner and Reasoner calls; every Perceiver invocation,
including multiple invocations within one plan step; all stage-aligned
Verifier calls and any upstream rechecks triggered by the controller;
and all failed calls and retries. Failed or retried invocations are
therefore charged separately, and a direct-return instance contributes
its native call. Deterministic spatial-tool executions that do not
invoke the base MLLM are not counted as model calls.  

For Tables~\ref{tab:exp-main-vsi}, \ref{tab:exp-main-sti}, and~\ref{tab:ablation}, each setting is evaluated in three runs on the same evaluation split under an identical configuration. We compute each task score and the benchmark overall score from the underlying per-question results separately for each run. Each run-level percentage is rounded to two decimal places before averaging across the three runs, and the resulting mean is again rounded to two decimal places for presentation. Overall scores are computed using the benchmark aggregation rule within each run and parenthesized differences are calculated from the displayed mean scores.

\subsection{Benchmarks and Metrics}
\label{app:datasets}

\paragraph{VSI-Bench-Debiased.}
We use the released VSI-Bench-Debiased v1~\citep{yang2025thinking,brown2025benchmarkdesignerstraintest} subset with 2,362 questions on indoor videos from ScanNet~\citep{dai2017scannet}, ScanNet++~\citep{yeshwanth2023scannet++}, and ARKitScenes~\citep{baruch2021arkitscenes}. The subset applies question-specific manual filtering to reduce cases solvable from answer priors or prior knowledge without inspecting the video. Its eight tasks comprise four numerical tasks: object counting, absolute distance, object size, and room size; and four multiple-choice tasks: relative distance, relative direction, appearance order, and route planning.

Numerical tasks use mean relative accuracy (MRA). For each prediction, relative accuracy is evaluated at ten relative-error thresholds from 5\% to 50\% in 5\% increments; MRA is the fraction of satisfied thresholds, averaged over questions. Multiple-choice tasks use answer accuracy. Relative-direction accuracy is first macro-averaged over the easy, medium, and hard subsets. The overall VSI-Bench-Debiased score is the unweighted mean of the eight task scores.

\paragraph{STI-Bench.}
We use the released STI-Bench test set, which contains 2,064 questions with five answer choices each, drawn from Omni6DPose~\citep{zhang2024omni6dpose},
ScanNet~\citep{dai2017scannet}, and Waymo~\citep{sun2020scalability}. It
covers three static tasks (dimensional measurement, spatial relation, and 3D
video grounding) and five dynamic tasks (displacement and path length, speed
and acceleration, egocentric orientation, trajectory description, and pose
estimation). Questions are generated from annotations in the source datasets
and reviewed by human annotators to exclude inaccurate descriptions and cases
with insufficient visual evidence. Answer choices reflect the measurement
precision of each environment. We report accuracy for each task and overall
accuracy across the full test set, equivalent to the mean task accuracy
weighted by sample count.

\section{Spatial Tools and Operation Interface}
\label{app:tool_definitions}

The operation interface $\mathcal{T}$ exposes the spatial primitives used by residual planning and targeted perception. Each operation consumes sampled-video references or outputs from preceding steps and writes a traceable raw record that can be cited by an evidence record. Operations that combine 3D quantities require a common coordinate frame. Unless external calibration is available, reconstructed geometry has unknown metric scale and gravity; distances therefore remain in reconstruction units, and planar directions are interpreted as horizontal only when the supplied up direction is supported by evidence.

\subsection{Visual Evidence Acquisition and 3D Localization}

\begin{toolcard}{Frame Inspection (\texttt{inspect\_frames})}
\toolfield{Input:}
\texttt{frame\_ids}: selected frame IDs from the sampled video.
\toolfield{Output:}
\texttt{frame\_ids} and \texttt{source=V}; the selected images are then provided to the Perceiver.
\toolfield{Description:}
Retrieves selected frames for direct visual inspection. The operation supports localized checks of visibility, spatial relations, and temporal order without resampling the original video.
\end{toolcard}

\begin{toolcard}{Object Detection (\texttt{object\_detection})}
\toolfield{Input:}
\texttt{objects}: at most eight requested object descriptions per call; \texttt{frame\_ids}: selected frame IDs. The optional \texttt{box\_threshold} and \texttt{text\_threshold} default to $0.35$ and $0.25$, respectively, with both constrained to $[0,1]$.
\toolfield{Output:}
Each detection records \texttt{frame\_id}, \texttt{object}, \texttt{box}, and \texttt{score}, where \texttt{box}$=[x_1,y_1,x_2,y_2]$ uses original-image pixel coordinates. The output also reports \texttt{earliest\_detection\_frame} over the queried frames.
\toolfield{Description:}
Applies GroundingDINO to each requested object description on the selected frames. A missed detection does not establish absence, and detections in different frames do not by themselves establish instance identity.
\end{toolcard}

\begin{toolcard}{Geometry Reconstruction (\texttt{reconstruct\_geometry})}
\toolfield{Input:}
\texttt{frame\_ids}: at most 16 frame IDs.
\toolfield{Output:}
\texttt{artifact}, \texttt{artifact\_sha256}, \texttt{coordinate\_frame}, \texttt{camera\_extrinsics}, and \texttt{camera\_intrinsics}. The saved NumPy archive stores \texttt{world\_points}, \texttt{world\_points\_conf}, camera parameters, frame IDs, and original image sizes. The interface marks both \texttt{metric\_scale} and \texttt{gravity} as \texttt{unknown}.
\toolfield{Description:}
Applies VGGT to estimate dense 3D points and camera parameters. Preprocessing pads images without cropping their original content.
\end{toolcard}

\begin{toolcard}{Object Localization (\texttt{localize\_objects})}
\toolfield{Input:}
\texttt{detection\_step} and \texttt{geometry\_step}: references to the detection and geometry steps, using their shared frames.
\toolfield{Output:}
Each entry in \texttt{locations} contains the coordinate-wise median \texttt{center}, \texttt{extent\_10\_90} (the 90th--10th percentile span on each axis), and \texttt{valid\_points}; \texttt{centers} provides the corresponding center list. All outputs retain the reconstruction coordinate frame.
\toolfield{Description:}
Lifts 2D detections into the reconstruction. For each mapped box, it retains valid 3D points with confidence at least the within-box median and discards boxes with fewer than ten retained points. Because localization is box-based, the retained points can include background, and repeated detections of one object remain separate unless explicitly associated downstream.
\end{toolcard}

\subsection{Distance Operations}

\begin{toolcard}{Euclidean Distance (\texttt{compute\_distance})}
\toolfield{Input:}
\texttt{a} and \texttt{b}: two 3D vector references that must share a coordinate frame.
\toolfield{Output:}
\texttt{distance} and \texttt{units=input coordinate units}.
\toolfield{Description:}
Resolves the two references and computes
\[
 d(a,b)=\lVert a-b\rVert_2.
\]
Without metric calibration, the result remains in reconstruction units.
\end{toolcard}

\begin{toolcard}{Closest-Point Distance (\texttt{compute\_closest\_point\_distance})}
\toolfield{Input:}
\texttt{object\_a} and \texttt{object\_b}: localized centers and extents in a shared coordinate frame; centers alone are also supported.
\toolfield{Output:}
\texttt{distance}, \texttt{distance\_type}, \texttt{closest\_point\_a}, \texttt{closest\_point\_b}, and \texttt{coordinate\_frame}. The center-only fallback sets \texttt{distance\_type=center\_approximation}.
\toolfield{Description:}
Estimates inter-object separation. When valid extents are available, it computes the minimum distance between the corresponding axis-aligned bounds. If only centers are available, it falls back to center distance. No metric conversion is applied to uncalibrated coordinates.
\end{toolcard}

\subsection{Coordinate Projections and Relative Direction}

\begin{toolcard}{Planar Projection (\texttt{project\_to\_bev})}
\toolfield{Input:}
\texttt{points}: referenced 3D points; \texttt{up}: a nonzero vector.
\toolfield{Output:}
The projected 2D \texttt{points} and \texttt{basis}.
\toolfield{Description:}
Projects the points onto the plane orthogonal to \texttt{up}. After normalizing \texttt{up}, the operation constructs an orthonormal basis for that plane. The projection represents a horizontal bird's-eye plane only when \texttt{up} is supported as the gravity direction; the operation neither estimates gravity nor renders an image.
\end{toolcard}

\begin{toolcard}{Observer Pose (\texttt{observer\_pose})}
\toolfield{Input:}
\texttt{observer}, \texttt{facing}, \texttt{locations}, and \texttt{up}. The two localized references are supplied through \texttt{locations}: \texttt{observer} identifies the object beside the observer, and \texttt{facing} identifies the object toward which the observer is oriented; \texttt{up} supplies the up direction.
\toolfield{Output:}
\texttt{observer\_position}, \texttt{forward\_vector}, \texttt{right\_vector}, and \texttt{up\_vector}.
\toolfield{Description:}
Constructs an observer-centered basis from the two localized references. The forward direction points from the observer reference to the facing reference, and the right direction is derived from the forward and up vectors. It does not infer observer height, body pose, or gravity beyond the supplied evidence.
\end{toolcard}

\begin{toolcard}{Observer-Frame Projection (\texttt{project\_to\_observer})}
\toolfield{Input:}
\texttt{points}, \texttt{observer\_position}, \texttt{forward\_vector}, \texttt{right\_vector}, and \texttt{up\_vector}: 3D points, the observer position, and the observer basis vectors.
\toolfield{Output:}
\texttt{observer\_coordinates} and the basis vectors used for the transformation.
\toolfield{Description:}
Expresses 3D points in the observer basis. For a point $\mathbf{p}$ and observer position $\mathbf{o}$, let $\mathbf{r}=\mathbf{p}-\mathbf{o}$. The returned coordinates are
\[
\begin{aligned}
 x_{\mathrm{right}} &= \mathbf{r}^{\top}\mathbf{right}, &
 y_{\mathrm{forward}} &= \mathbf{r}^{\top}\mathbf{forward}, &
 z_{\mathrm{up}} &= \mathbf{r}^{\top}\mathbf{up}.
\end{aligned}
\]
\end{toolcard}

\begin{toolcard}{Relative-Direction Classification (\texttt{compute\_rel\_dir})}
\toolfield{Input:}
\texttt{target}, \texttt{observer}, \texttt{forward}, \texttt{right}, \texttt{up}, and \texttt{back\_angle}: the 3D target, observer reference, basis vectors, and back threshold. The default back threshold is $135^\circ$.
\toolfield{Output:}
\texttt{direction}, \texttt{right\_component}, \texttt{forward\_component}, and \texttt{turn\_angle\_deg}.
\toolfield{Description:}
Projects a 3D target into the observer frame. Let $\theta$ be the angle between the forward axis and the target direction. The target is classified as \texttt{back} when $\theta\geq\texttt{back\_angle}$; otherwise, the sign of the right component determines \texttt{left} or \texttt{right}.
\end{toolcard}

\begin{toolcard}{Observer-Coordinate Direction Classification (\texttt{classify\_dir})}
\toolfield{Input:}
\texttt{observer}: coordinates already expressed in the observer frame; \texttt{back\_angle}: the back threshold, with a default of $135^\circ$.
\toolfield{Output:}
\texttt{left}, \texttt{right}, or \texttt{back}; \texttt{unknown} if no supported forward axis is available.
\toolfield{Description:}
Applies the same classification rule as \texttt{compute\_rel\_dir} to coordinates already expressed in the observer frame, without assuming an image- or world-axis orientation when no supported forward axis is available.
\end{toolcard}

\subsection{Spatial Grouping and Cross-Frame Association}

\begin{toolcard}{Spatial Clustering (\texttt{cluster\_detections})}
\toolfield{Input:}
\texttt{points}: 3D points; \texttt{radius}: a positive radius in the same coordinate units as the points.
\toolfield{Output:}
\texttt{clusters}, \texttt{count}, and \texttt{radius}; cluster entries are zero-based indices into the input list.
\toolfield{Description:}
Builds a graph over the points, connecting pairs whose Euclidean distance is at most \texttt{radius}, and returns its connected components as \texttt{clusters}. Because connectivity is transitive, a component can span more than \texttt{radius}, and its count should not be interpreted directly as an object count.
\end{toolcard}

\begin{toolcard}{Cross-Frame Instance Association (\texttt{associate\_instance})}
\toolfield{Input:}
\texttt{locations}: localized detections in a shared coordinate frame; \texttt{radius}: the center-distance grouping radius.
\toolfield{Output:}
Each entry in \texttt{instances} contains \texttt{instance\_id}, \texttt{member\_indices}, \texttt{frame\_ids}, \texttt{cluster\_center}, and \texttt{association\_radius}.
\toolfield{Description:}
Groups localized detections with the same object category whose centers fall within \texttt{radius}. The association is a geometric grouping cue rather than an identity guarantee; ambiguous category labels, coordinate alignment, or temporal continuity remain unresolved evidence conditions.
\end{toolcard}

\subsection{Composite Operations and Step Accounting}
\label{app:composite-operations}

The operation interface provides registered primitive operations.
The Planner may invoke a primitive directly or combine multiple
primitives into a composite operation according to the requirements
of the task.
A composite operation is constructed during planning rather than
selected from a fixed library of predefined compositions.
For each internal operation, the Planner specifies its arguments,
input references, dependencies, and documented output fields.
Internal operations execute in dependency order and retain separate
raw records for verification.
The composite also declares the outputs available to subsequent
plan steps. Each entry in the plan counts as one step, whether it invokes
a single primitive or a composite operation.
We encourage the Planner to use at most five steps per plan,
but impose no fixed upper bound.

\section{Discussion and Limitations}
\label{app:discussion}

\subsection{Implications for Residual Test-Time Computation}

Prior work shows that the effectiveness
of test-time computation depends on problem difficulty, motivating
adaptive computation strategies~\citep{snell2025scaling}.
Our analyses examine a related allocation question in video spatial
reasoning: which unresolved requirements need additional computation,
and which intermediate states need to be recomputed after a failure. On the fixed routing subsets, forcing supplementary acquisition for direct-return instances and suppressing refinement for refinement instances both reduce performance across the three base models (Fig.~\ref{fig:exp-routing}). Test-time scaling is allocation-sensitive: useful computation is determined by what remains unresolved for the current model--instance pair, rather than by a fixed inference depth. The recovery study shows the same pattern after refinement begins. Selective recovery reuses compatible plans and evidence, reducing repeated model calls relative to full restart, while larger recovery budgets do not yield monotonic gains (Fig.~\ref{fig:correction-efficiency}). These results support residual computation as an \emph{allocation principle}: acquire only missing evidence and recompute only states invalidated by a localized failure.

\subsection{Scope and Extensions}
\label{app:limitations}

ReTAC's practical scope is set primarily by the quality of its state estimates and supplementary operations. A native call that incorrectly marks incomplete evidence as supported may terminate early, and using the same base MLLM for planning, perception, reasoning, and verification can preserve correlated errors across roles. Its supplementary evidence is likewise bounded by the operation interface; for example, the current VGGT-based geometry does not provide metric scale, consistent with the less uniform gains on metric tasks. More calibrated support estimates, independent verification signals, and uncertainty-aware perception outputs could strengthen routing and recovery without changing the residual formulation.

The empirical study covers three heterogeneous MLLMs and two video spatial reasoning benchmarks, with mean model calls as the primary compute proxy. We plan to extend the evaluation to
  other multimodal tasks to assess the broader applicability of
  residual allocation, and to measure token usage and tool latency
  for a more complete assessment of efficiency. We also plan to
  conduct targeted interventions to isolate the contributions of
  native state conditioning, evidence reuse, verification structure,
  and cached verification results under different computation budgets.
  Our current evaluation uses the original STI-Bench release.
  Given the coordinate and timestamp annotation issues reported
  by \citet{sun2026restisourcegroundedauditrepair}, these results
  should be interpreted within the limitations of that release.

\section{Agent and Ablation Prompts}
\label{app:agent-prompts}

This appendix gives the role-specific prompts used by ReTAC and the two component-replacement ablations. Shared constraints are factored into prefixes: \texttt{NATIVE}, \texttt{PLANNER}, \texttt{PERCEIVER}, and \texttt{REASONER} prepend \texttt{COMMON}; \texttt{CHECK\_COMMON} prepends \texttt{COMMON} to the shared verification instructions; each stage verifier then prepends \texttt{CHECK\_COMMON}. All prompts are fixed across base models.

\subsection{Shared Prompt Conventions}
\label{app:prompt-common}

The shared prefix standardizes output format, identifiers, evidence bookkeeping, and answer representation across roles.

\begin{agentprompt}{COMMON}
Return compact JSON matching the supplied schema, using actual findings rather than placeholders. Keep each explanation to one short sentence. Merge observations only when fact, object identity, time, and source agree; keep distinct or conflicting evidence separate.

Use unique IDs: r for requirements, e for evidence, s for steps, and z for reasoning claims. Frame IDs are ONE-BASED sampled-video positions; reference only supplied frames.

candidate_answer is a choice letter or one number encoded as a string in the requested units, or JSON null when unavailable. Never output the literal string "null". State uncertainty without inventing facts.
\end{agentprompt}

\subsection{Native State Elicitation and Residual Routing}
\label{app:prompt-native}

The native prompt produces the requirement--evidence state used to decide between direct return and residual refinement.

\begin{agentprompt}{NATIVE}
Answer directly from the supplied video using native visual and spatial reasoning; no supplementary tools are available. Identify the concrete facts needed for the answer(e.g., identity, count, position, relation, dimension, or time) and record the corresponding native observations.

Label a requirement SUPPORTED only when the observations resolve it; otherwise label it UNRESOLVED and state the missing information. Requirements describe facts, not the final answer task. Every SUPPORTED requirement must reference matching evidence through requirement_id and evidence_ids. unresolved_requirements contains exactly the UNRESOLVED IDs. Preserve object identity, time, and spatial reference frame.

Native evidence uses method native, acquisition_step native_N0, and null tool-source fields. Put usable numerical quantities in claim.value and units, basis, and uncertainty in claim.text. Physical estimates based on familiar-object sizes are assumptions, not measurements. Return the best-supported candidate without duplicating the same observation across frames.
\end{agentprompt}

\subsection{Residual Planning, Perception, and Reasoning}
\label{app:prompt-residual}

On the refinement branch, the three roles acquire only missing evidence, register it with provenance, and reason over the resulting active evidence set.

\paragraph{Residual Planner.}
The Planner converts unresolved requirements into an executable plan while preserving supported native state.

\begin{agentprompt}{PLANNER}
Plan residual acquisition while preserving native requirements and observations. Retain all requirements and add any relevant requirement omitted by the native state. Every requirement targeted by supplementary acquisition, including additions during REPLAN, must appear in unresolved_requirements.

A plan step may invoke one registered primitive operation or a composition of registered primitives that you construct for the task. For each internal operation, declare its arguments, input references, dependencies, and documented output fields. Declare the outputs exposed by the composite and preserve a separate raw record for each internal operation.

Prefer plans with at most five steps. This is a recommendation, not a hard limit. If more than five steps are needed, review the plan for redundant acquisitions, unnecessary intermediate operations, or an unsuitable solution strategy. Revise the plan where appropriate and retain additional steps when they are necessary.

Use only registered primitives and documented output fields. Reference V, evidence IDs, or preceding operation outputs as inputs and declare dependencies. Measured arguments use {ref, field}.

object_detection takes category descriptions, not instance IDs. reconstruct_geometry uses at most 16 frames. localize_objects requires outputs from preceding object_detection and reconstruct_geometry operations, with both dependencies declared. These operations may appear in separate steps or within a composite step, provided that their dependency order is preserved. inspect_frames is not a detector.

Detection boxes are 4D pixel bounds; localization exposes 3D centers at locations.0.center. Reconstruction returns an artifact and camera matrices, not object positions, metric scale, or gravity. Never invent output paths. A missed detection does not establish absence. Use targeted frame inspection when sufficient.

On REPLAN, repair the diagnosed gap by changing the affected acquisition steps. Changing only expected_output or repeating an unchanged failed plan is not a repair.
\end{agentprompt}

\paragraph{Targeted Perceiver.}
The Perceiver turns one planned acquisition into source-linked evidence rather than a final task answer.

\begin{agentprompt}{PERCEIVER}
Interpret only the current acquisition, its input records, and source frames. Produce observations for the target requirements, not the final answer. Cite raw tool-record IDs, not step IDs. For a tool-extracted value, set tool_field to the actual output-relative path and claim.value to that value. Direct visual claims use tool_field null and may use tool_record_id null. Give numerical quantities explicit units and a short basis in claim.text.

Preserve object identity, time, coordinate frame, and scale; report unresolved requirements and conflicts.

A failed tool is not evidence. Detection alone establishes neither unique identity nor absence; first-appearance claims require checking preceding sampled frames. VGGT has unknown metric scale and gravity, so metric conversion or a horizontal reference requires supporting evidence. Approximate physical estimates must state the visible cue and assumed reference size.
\end{agentprompt}

\paragraph{Evidence-Grounded Reasoner.}
The Reasoner derives the candidate from active evidence without introducing new visual observations.

\begin{agentprompt}{REASONER}
Reason only from active evidence. Each claim's support_ids must reference active evidence or earlier supported claims. Track identity, time, coordinates, units, and conflicts. Return the best-supported candidate and retain unresolved requirements or conflicts.

For arithmetic, provide an operator, traced operands, and result; semantic claims use calculation null. Evidence operands reference claim.value, computed operands reference calculation.result, and every operand appears in support_ids. Do not create a numeric claim.value for a semantic conclusion solely to make it referenceable; cite the underlying numerical evidence instead.

final_conclusion and candidate_answer must agree on the option or quantity. Put uncertainty in the reasoning, not an answer range. Retain estimates supported by explicit evidence and assumptions when exact calibration is unavailable.
\end{agentprompt}

\subsection{Stage-Aligned Verification}
\label{app:prompt-verification}

Verification mirrors the planning--perception--reasoning decomposition; each stage receives only the inputs needed for its criteria in Appendix~\ref{app:verification-details}.

\paragraph{Shared Verification Instructions.}
The shared verifier prefix defines verdict semantics, recovery actions, and targeted upstream rechecks.

\begin{agentprompt}{CHECK\_COMMON}
Evaluate every named criterion independently: PASS requires support, FAIL a definite error, and UNKNOWN insufficient support to decide. Identify specific targets. For FAIL or UNKNOWN, propose a concrete supported correction when available; otherwise set action to null and revision to empty.

REPERCEIVE repeats the existing operation, arguments, inputs, dependencies, and frame selection. Changed acquisition requires REPLAN; REREASON changes reasoning only. Recovery target lists contain IDs.

If a later check raises a specific concern about an earlier PASS, report upstream_concern with stage, criterion, target, reason, and revision. The controller rechecks that criterion before routing recovery; do not overturn a PASS on a hypothesis.
\end{agentprompt}

\paragraph{Planning Verifier.}
The planning verifier checks plan completeness and feasibility without seeing observations, reasoning, or the candidate answer.

\begin{agentprompt}{PLAN\_CHECK}
You receive the question, options, plan, tool interfaces, and reference metadata, but NO video, observations, reasoning, or answer.

plan_completeness: Check whether the plan covers the entities, relations, and references needed for the answer; could an omitted fact change the answer while planned facts remain fixed?

operation_feasibility: Check primitive operations, arguments, input references, output capability, and dependency order both within composite steps and across plan steps. Native-supported requirements need no reacquisition. inspect_frames can acquire spatial or temporal observations, not only object presence. Judge what execution COULD supply; withheld observations are not a planning failure. Metric quantities require a feasible calibration or estimation procedure.

A supported planning failure proposes REPLAN; without a concrete supported revision, stop unverified.
\end{agentprompt}

\paragraph{Perception Verifier.}
The perception verifier validates every active evidence record against its cited source before checking requirement coverage.

\begin{agentprompt}{PERCEPTION\_CHECK}
Compare all active native and supplementary evidence with the cited frames and raw operation records. Check source_validity and requirement_coverage separately. Preserve object identity, time, coordinate frame, and scale.

source_validity: Check every active evidence record. Validate its frame references, operation records, and cited output fields, then determine whether the source supports the observation. Return PASS when support is established, FAIL for a definite inconsistency, and UNKNOWN when the source is inaccessible or support is ambiguous. Include an individual check for every active evidence ID in the observations field defined by the supplied schema.

requirement_coverage: Check whether each requirement is resolved by valid active evidence. Return PASS for a resolved requirement. Return FAIL when the required evidence is missing or contradicted. Return UNKNOWN when support or relevance remains uncertain. A record whose source validity is UNKNOWN does not by itself establish either coverage or a definite coverage failure. Consider any other active evidence for the same requirement.

Propose REPERCEIVE when repeating the affected acquisition with unchanged operations, arguments, inputs, dependencies, and frame selection can repair the evidence. If the acquisition must change, or a native error has no corresponding acquisition step, report a targeted planning concern with a concrete proposed revision. The controller rechecks the implicated planning criterion before selecting REPLAN. If no concrete supported revision is available, set action to null and revision to empty.
\end{agentprompt}

\paragraph{Reasoning Verifier.}
The reasoning verifier checks the evidence-to-answer derivation after planning and perception pass.

\begin{agentprompt}{REASON\_CHECK}
Check evidence_support, inference_validity, and answer_consistency along evidence -> cited claims -> conclusion -> answer. Validate premises, semantic or geometric inferences, and calculations. Map conclusions to option IDs and check units and requested rounding.

A plausible answer does not excuse unsupported premises. A specific evidence gap first raises a perception concern; if the plan and evidence are valid, reasoning, calculation, mapping, or formatting errors propose REREASON.
\end{agentprompt}

\subsection{Component-Replacement Ablation Prompts}

These prompts implement the component replacements in Sections~\ref{sec:ablation} and~\ref{sec:reflection-analysis}: \texttt{GENERAL\_PLANNING} removes native-state conditioning, while \texttt{GENERAL\_REFLECTION} replaces stage-aligned verification and localized recovery with a trajectory-level decision.

\paragraph{General Planning.}
\label{app:general-planning-prompt}
The general planner decides whether to use supplementary computation without access to native support labels, native evidence, or native unresolved requirements.

\begin{ablationprompt}{GENERAL\_PLANNING}
You are a general adaptive planner for video spatial reasoning. You receive the video, question, answer options, operation interfaces, and descriptions of the facts needed to answer. Decide whether supplementary computation is useful from these inputs.

Return RETURN_NATIVE when direct reasoning from the input appears sufficient. The controller then returns the stored native candidate unchanged; do not generate a new answer. You are not given that candidate, native support labels, native observations, or native unresolved requirements.

Otherwise return SUPPLEMENT with a complete plan covering all facts needed for the answer. Retain supplied requirement IDs and add missing requirements. Assume no requirement is already supported: every downstream fact must be established by the new acquisition because native observations are unavailable.

Use ordered steps, including prerequisites, with registered operations and documented arguments and outputs. Each step specifies target_requirement, operation, arguments, input_references, dependencies, and expected_output. Inputs may reference V or earlier steps, not native evidence. Use {ref, field} for prior outputs, declare dependencies, and never invent output fields. Targeted frame inspection is available. Detection alone establishes neither identity nor absence; reconstructed geometry has no metric scale or gravity.

Return compact JSON with exactly these top-level fields:
- decision: RETURN_NATIVE or SUPPLEMENT.
- rationale: one short sentence explaining the computation decision.
- plan: null for RETURN_NATIVE; otherwise an object with requirements and steps.

For SUPPLEMENT, address the full task rather than a presumed subset of native failures. Describe acquisition, not anticipated observations or an answer. Reference only supplied sampled-frame IDs.
\end{ablationprompt}

\paragraph{General Reflection.}
\label{app:general-reflection-prompt}
The general reflector makes one accept-or-retry decision over the full trajectory instead of localizing a failure to a specific stage.

\begin{ablationprompt}{GENERAL\_REFLECTION}
You are a general reflector for video spatial reasoning. Review the current candidate and the trajectory as a whole using the question, options, fixed native output, current plan, active evidence, reasoning, supplied video sources, raw tool records, and previous feedback. Decide whether the current answer is sufficiently supported.

Return ACCEPT when the answer is adequately supported. Otherwise return RETRY with brief, concrete feedback on what the next attempt should address. Do not assume unseen source content or treat previous feedback as new evidence.

Use one overall decision. Do not output per-stage verdicts, criterion scores, structured correction targets, or stage-specific action codes. Do not rewrite the candidate. ACCEPT returns it unchanged. RETRY reruns supplementary planning, perception, and reasoning from the fixed native attempt using accumulated feedback; the controller enforces the retry budget.

Return only compact JSON with exactly two fields:
- decision: the literal string ACCEPT or RETRY.
- feedback: a brief explanation, at most two sentences.
\end{ablationprompt}

\section{Worked Example: Perception Error and Selective Recovery}
\label{app:worked-example}

We trace a refinement trajectory in which the residual Perceiver produces a source-inconsistent temporal observation. The example follows the state definitions, operation interface, and prompts in Appendices~\ref{app:method-details}--\ref{app:agent-prompts}. Unlike Case~2 in Fig.~\ref{fig:case-study}, which corrects a reasoning error with \texttt{REREASON}, this trajectory fails at perception and triggers \texttt{REPERCEIVE}. YAML records are abbreviated, with model outputs, raw operation records, and controller updates separated.

\subsection{Example Input and Conventions}

We use an appearance-order query with directly traceable temporal evidence. Let $x=(V,q,\mathcal{O})$, where $V$ contains 32 ordered sampled frames, $q=$ ``Which object appears first in the provided frame sequence: a table, a lamp, or a stool?'', and $\mathcal{O}=\{\mathrm{A}:\text{table},\mathrm{B}:\text{lamp},\mathrm{C}:\text{stool}\}$. The corresponding instances are \texttt{table\_1}, \texttt{lamp\_1}, and \texttt{stool\_1}, whose first appearances in this trace are sampled frames 8, 24, and 4. Requirements \texttt{r\_table}, \texttt{r\_lamp}, and \texttt{r\_stool} encode these facts. Frame indices are one-based, ranges are inclusive, and temporal claims refer only to the sampled sequence. The recovery budget is $T=2$.

\newif\iftracefirstsection
\newtcolorbox{tracebox}[1]{
    enhanced,
    breakable,
    colback=black!5,
    colframe=black,
    colbacktitle=black!65,
    coltitle=white,
    boxrule=0.4pt,
    arc=0pt,
    left=7pt,right=7pt,top=6pt,bottom=6pt,
    toptitle=3pt,bottomtitle=3pt,
    fonttitle=\appendixboxtitlefont,
    fontupper=\footnotesize,
    before upper={\tracefirstsectiontrue},
    title={#1}
  }
\newcommand{\traceheading}[1]{%
  \par\iftracefirstsection
    \tracefirstsectionfalse
  \else
    \smallskip\noindent
    {\color{black!55}\rule{\linewidth}{0.35pt}}\par\nobreak
  \fi
  \smallskip\noindent\textbf{\textit{#1}}\par\nobreak\noindent\ignorespaces
}

\subsection{Native State and Initial Residual Refinement}

The native state resolves the table and lamp appearances but leaves the stool appearance unresolved. Residual refinement therefore targets only \texttt{r\_stool}.

\begin{tracebox}{1. Native state and routing}

\traceheading{Input.}
The native call receives only $x$ and uses base model $M$, without tool descriptions or tool outputs. Its producing-step identifier is \texttt{native\_N0}.

\traceheading{Model output.}
The call returns
$S^{\mathrm{nat}}=(\mathcal{R}^{\mathrm{nat}},\mathcal{E}^{\mathrm{nat}},\mathcal{U}^{\mathrm{nat}},\hat{y}^{\mathrm{nat}})$:
\begin{verbatim}
requirements:
  - id: r_table
    object_id: table_1
    status: SUPPORTED
    evidence_ids: [e_table]
  - id: r_lamp
    object_id: lamp_1
    status: SUPPORTED
    evidence_ids: [e_lamp]
  - id: r_stool
    object_id: stool_1
    status: UNRESOLVED
    evidence_ids: []
    reason: Cannot establish the first appearance.
evidence:
  - id: e_table
    requirement_id: r_table
    object_id: table_1
    claim:
      value: 8
      text: table_1 first appears at sampled frame 8.
    sources:
      video: V
      frame_range: [1, 8]
      tool_record_id: null
      tool_field: null
    method: native
    acquisition_step: native_N0
    input_record_ids: []
  - id: e_lamp
    requirement_id: r_lamp
    object_id: lamp_1
    claim:
      value: 24
      text: lamp_1 first appears at sampled frame 24.
    sources:
      video: V
      frame_range: [1, 24]
      tool_record_id: null
      tool_field: null
    method: native
    acquisition_step: native_N0
    input_record_ids: []
unresolved_requirements: [r_stool]
candidate_answer: null
\end{verbatim}

\traceheading{Controller update.}
The labels give $\mathcal{U}^{\mathrm{nat}}=\{\texttt{r\_stool}\}$, so the instance enters refinement. Because no native answer is established, $\hat{y}^{\mathrm{nat}}=\varnothing$ is encoded as \texttt{null}. An empty $\mathcal{U}^{\mathrm{nat}}$ would instead return the native candidate with status \texttt{DIRECT}.

\end{tracebox}

\begin{tracebox}{2. Residual plan}

\traceheading{Input.}
The Planner receives $(x,S^{\mathrm{nat}},\mathcal{T})$. Here \texttt{object\_detection(objects, frame\_ids)} returns frame-indexed detections and \texttt{earliest\_detection\_frame}; preceding sampled frames must still be inspected because a missed detection does not establish absence.

\traceheading{Model output.}
The Planner returns $P_0=(\mathcal{S}_0,\mathcal{R}_0,\mathcal{U}_0)$:
\begin{verbatim}
requirements: [r_table, r_lamp, r_stool]
unresolved_requirements: [r_stool]
steps:
  - id: s_stool
    target_requirement: [r_stool]
    operation: object_detection
    arguments:
      objects: [stool]
      frame_ids: [1, 2, 3, 4, 5, 6, 7, 8, 9, 10, 11, 12]
    input_references: [V]
    dependencies: []
    expected_output: >
      Frame-indexed stool detections within sampled frames 1--12.
      Use the detections and source frames to establish the first
      appearance if supported; otherwise keep r_stool unresolved.
\end{verbatim}

\traceheading{Controller update.}
No requirement is added, so $\mathcal{R}_0=\mathcal{R}^{\mathrm{nat}}$ and $\mathcal{U}_0=\mathcal{U}^{\mathrm{nat}}$. Only \texttt{r\_stool} is reacquired. Source support is decided after execution; changing the queried frames would alter the plan and require \texttt{REPLAN}.

\end{tracebox}

\begin{tracebox}{3. Supplementary perception and evidence merge}

\traceheading{Input.}
The Perceiver executes \texttt{s\_stool}, retains the raw record, and interprets it with the queried source frames.

\traceheading{Raw operation record.}
\begin{verbatim}
tool_record:
  id: D0
  acquisition_step: s_stool
  video: V
  frame_ids: [1, 2, 3, 4, 5, 6, 7, 8, 9, 10, 11, 12]
  earliest_detection_frame: 4
\end{verbatim}

\traceheading{Perceiver output.}
The Perceiver incorrectly records frame 10 despite citing a raw field whose value is 4:
\begin{verbatim}
new_evidence:
  - id: e_stool_v0
    requirement_id: r_stool
    object_id: stool_1
    claim:
      value: 10
      text: stool_1 first appears at sampled frame 10.
    sources:
      tool_record_id: D0
      tool_field: earliest_detection_frame
      video: V
      frame_range: [1, 12]
    method: object_detection_with_source_inspection
    acquisition_step: s_stool
    input_record_ids: [D0]
acquisition_failures: []
reported_unresolved_requirements: []
reported_conflicts: []
\end{verbatim}

\traceheading{Controller update.}
The active evidence becomes
\[
\mathcal{E}_0=
\{\texttt{e\_table},\texttt{e\_lamp},\texttt{e\_stool\_v0}\}.
\]
The raw record \texttt{D0} remains in the execution history. Empty unresolved/conflict lists do not validate the observation; verification checks source validity and requirement coverage separately.

\end{tracebox}

\begin{tracebox}{4. Initial reasoning}

\traceheading{Input.}
The Reasoner receives $(q,\mathcal{O},P_0,\mathcal{E}_0)$.

\traceheading{Model output.}
It returns $(Z_0,\hat{y}_0)$:
\begin{verbatim}
reasoning:
  - id: z_table_before_stool_v0
    support_ids: [e_table, e_stool_v0]
    calculation:
      operator: "<"
      operands:
        - {ref: e_table, field: claim.value}
        - {ref: e_stool_v0, field: claim.value}
      result: true
    conclusion: The table appears before the stool.
  - id: z_table_before_lamp_v0
    support_ids: [e_table, e_lamp]
    calculation:
      operator: "<"
      operands:
        - {ref: e_table, field: claim.value}
        - {ref: e_lamp, field: claim.value}
      result: true
    conclusion: The table appears before the lamp.
  - id: z_answer_v0
    support_ids: [z_table_before_stool_v0,
                  z_table_before_lamp_v0]
    calculation: null
    conclusion: "A: table"
candidate_answer: A
unresolved_requirements: []
unresolved_conflicts: []
\end{verbatim}

\traceheading{Controller update.}
Set $t=0$, $H=()$, and $C=\varnothing$. The comparisons are consistent with the active values, but the stool premise is source-inconsistent; the candidate remains provisional.

\end{tracebox}

\subsection{Stage-Aligned Verification and Error Localization}

Verification follows planning--perception--reasoning order. The plan passes, but the stool observation contradicts its cited source, so verification stops at perception.

\begin{tracebox}{5a. Planning verification}

\traceheading{Check input.}
The planning verifier receives only $(q,\mathcal{O},P_0,\mathcal{T})$, without the video, observation contents, reasoning trace, or candidate answer. Deterministic checks use the declared operation arguments, references, and dependencies.

\traceheading{Check output.}
\begin{verbatim}
planning:
  plan_completeness:
    result: PASS
    reason: >
      The three requirements cover the first appearance of each
      listed object under the same temporal reference.
  operation_feasibility:
    result: PASS
    rule_checks:
      registered_operation: PASS
      arguments: PASS
      input_references_and_dependencies: PASS
    reason: >
      object_detection can provide stool detections over the
      declared sampled frames; source support is checked later.
  result: PASS
  origin: fresh
\end{verbatim}

\traceheading{Controller update.}
Record $v_0^{\mathrm{plan}}=\texttt{PASS}$. Feasibility does not validate \texttt{e\_stool\_v0}.

\end{tracebox}

\begin{tracebox}{5b. Perception failure and recovery routing}

\traceheading{Check input.}
The perception verifier receives $P_0$, $\mathcal{E}_0$, the cited sampled frames, raw record \texttt{D0}, and the relevant interface definition. Native and supplementary observations are checked under the same source-validity criterion.

\traceheading{Check output.}
\begin{verbatim}
perception:
  source_validity:
    result: FAIL
    observations:
      e_table:
        result: PASS
        reason: Frame 8 shows table_1; frames 1--7 do not.
      e_lamp:
        result: PASS
        reason: Frame 24 shows lamp_1; frames 1--23 do not.
      e_stool_v0:
        result: FAIL
        rule_reason: >
          claim.value is 10, but D0.earliest_detection_frame is 4.
        source_reason: >
          Frame 4 shows stool_1 and contradicts a first
          appearance at frame 10.
  requirement_coverage:
    result: FAIL
    requirement_results:
      r_table: PASS
      r_lamp: PASS
      r_stool: FAIL
    reason: r_stool has no source-valid active observation.
  result: FAIL
  origin: fresh
reasoning:
  result: null
  origin: not_run
\end{verbatim}

Thus
$\boldsymbol{v}_0=(\texttt{PASS},\texttt{FAIL},\text{not run})$.
Reasoning verification is not executed after the perception failure; \texttt{not\_run} is distinct from \texttt{UNKNOWN}.

\traceheading{Diagnostic proposal $J_0$.}
\begin{verbatim}
target: [e_stool_v0]
diagnosis: >
  The observation misreports the cited earliest detection and
  ignores the source-supported occurrence at sampled frame 4.
revision: >
  Repeat s_stool with the same acquisition definition, inspect
  the cited detection and preceding sampled frames, replace the
  invalid stool observation, and regenerate dependent reasoning.
\end{verbatim}

\traceheading{Controller output.}
$\operatorname{Route}(\boldsymbol{v}_0,J_0)$ returns
$F_0=(a_0,\tau_0,\delta_0,\rho_0)$:
\begin{verbatim}
decision: REPERCEIVE
target: [e_stool_v0]
diagnosis: >
  e_stool_v0 is inconsistent with its cited raw field and frame.
revision: >
  Repeat s_stool with unchanged operation, arguments, inputs,
  and dependencies; correct the stool observation and regenerate
  the dependent reasoning and answer.
\end{verbatim}

\traceheading{Controller update.}
The same acquisition can repair the error, so no planning recheck is needed. Since $t=0<T$, $F_0$ is appended to $H$. Planning checks and source-validity checks for \texttt{e\_table} and \texttt{e\_lamp} remain cacheable; checks depending on \texttt{e\_stool\_v0} are invalidated.

\end{tracebox}

\subsection{Selective Recovery, Reverification, and Final Return}

Selective recovery replaces the invalid stool observation and recomputes its dependents while reusing the plan and unaffected native evidence.

\begin{tracebox}{6. Reperception and dependency update}

\traceheading{Recovery input.}
The recovery call receives $(x,M,\mathcal{T},P_0,\mathcal{E}_0,H,F_0)$ and keeps $P_1=P_0$. Step \texttt{s\_stool} therefore uses the same operation, frame IDs, inputs, and dependencies.

\traceheading{Raw operation record.}
Repeating the acquisition produces \texttt{D1}:
\begin{verbatim}
tool_record:
  id: D1
  acquisition_step: s_stool
  video: V
  frame_ids: [1, 2, 3, 4, 5, 6, 7, 8, 9, 10, 11, 12]
  earliest_detection_frame: 4
\end{verbatim}
The raw output is unchanged; the correction comes from the Perceiver's revised interpretation and source-frame inspection.

\traceheading{Corrected Perceiver output.}
\begin{verbatim}
new_evidence:
  - id: e_stool_v1
    requirement_id: r_stool
    object_id: stool_1
    claim:
      value: 4
      text: stool_1 first appears at sampled frame 4.
    sources:
      tool_record_id: D1
      tool_field: earliest_detection_frame
      video: V
      frame_range: [1, 4]
    method: object_detection_with_source_inspection
    acquisition_step: s_stool
    input_record_ids: [D1]
acquisition_failures: []
reported_unresolved_requirements: []
reported_conflicts: []
\end{verbatim}

\traceheading{Controller update.}
The revised active evidence is
\[
\mathcal{E}_1=
\{\texttt{e\_table},\texttt{e\_lamp},\texttt{e\_stool\_v1}\}.
\]
The source range $[1,4]$ supports the corrected claim, while the planned acquisition remains frames 1--12. \texttt{e\_stool\_v0} leaves the active set but remains in the history with \texttt{D0}; the two native observations remain active.

\traceheading{Reasoner output.}
Regenerating reasoning from $(q,\mathcal{O},P_1,\mathcal{E}_1;H,F_0)$ gives
$(Z_1,\hat{y}_1)$:
\begin{verbatim}
reasoning:
  - id: z_stool_before_table_v1
    support_ids: [e_stool_v1, e_table]
    calculation:
      operator: "<"
      operands:
        - {ref: e_stool_v1, field: claim.value}
        - {ref: e_table, field: claim.value}
      result: true
    conclusion: The stool appears before the table.
  - id: z_stool_before_lamp_v1
    support_ids: [e_stool_v1, e_lamp]
    calculation:
      operator: "<"
      operands:
        - {ref: e_stool_v1, field: claim.value}
        - {ref: e_lamp, field: claim.value}
      result: true
    conclusion: The stool appears before the lamp.
  - id: z_answer_v1
    support_ids: [z_stool_before_table_v1,
                  z_stool_before_lamp_v1]
    calculation: null
    conclusion: "C: stool"
candidate_answer: C
unresolved_requirements: []
unresolved_conflicts: []
\end{verbatim}

\traceheading{Cache update.}
Only checks whose dependencies remain unchanged are retained:
\begin{verbatim}
retained_checks:
  - planning.plan_completeness
  - planning.operation_feasibility
  - source_validity(e_table)
  - source_validity(e_lamp)
invalidated_checks:
  - source_validity(e_stool_v0)
  - perception.requirement_coverage
requires_verification:
  - source_validity(e_stool_v1)
  - perception.requirement_coverage
  - reasoning.evidence_support
  - reasoning.inference_validity
  - reasoning.answer_consistency
\end{verbatim}
No reasoning check is reusable because the initial verification stopped at perception. Set $t=1$ and verify the revised state.

\end{tracebox}

\begin{tracebox}{7. Reverification and return}

\traceheading{Input.}
The verifier receives $\Sigma_1=(P_1,\mathcal{E}_1,Z_1,\hat{y}_1)$ with the still-valid cache entries. The unchanged planning criteria remain \texttt{PASS}; perception first verifies the replacement observation and coverage before reasoning is checked.

\traceheading{Check output.}
\begin{verbatim}
planning:
  plan_completeness: {result: PASS, origin: cached}
  operation_feasibility: {result: PASS, origin: cached}
  result: PASS
perception:
  source_validity:
    result: PASS
    observations:
      e_table: {result: PASS, origin: cached}
      e_lamp: {result: PASS, origin: cached}
      e_stool_v1:
        result: PASS
        origin: fresh
        reason: >
          claim.value=4 matches D1.earliest_detection_frame;
          frame 4 shows stool_1 and frames 1--3 do not.
  requirement_coverage:
    result: PASS
    origin: fresh
    requirement_results:
      r_table: PASS
      r_lamp: PASS
      r_stool: PASS
  result: PASS
reasoning:
  evidence_support:
    result: PASS
    reason: >
      Every critical claim cites active evidence or an earlier
      supported claim, and each premise is source-supported.
  inference_validity:
    result: PASS
    reason: >
      Both 4 < 8 and 4 < 24 hold for the cited evidence values.
  answer_consistency:
    result: PASS
    reason: The final conclusion and candidate both map to C.
  result: PASS
  origin: fresh
\end{verbatim}

\traceheading{Controller output and return.}
Now
$\boldsymbol{v}_1=(\texttt{PASS},\texttt{PASS},\texttt{PASS})$:
\begin{verbatim}
decision: ACCEPT
target: null
diagnosis: All required checks pass.
revision: null
answer: C
status: VERIFIED
correction_rounds: 1
\end{verbatim}
The corrected state is accepted after one recovery action. The unused budget is not consumed; a state that passes at $t=T$ is still accepted because $T$ limits recovery actions, not the final verification.

\end{tracebox}

This trajectory isolates the recovery scope: the perception error invalidates the stool observation, requirement coverage, and dependent reasoning, while the plan and unrelated native evidence remain reusable.

\end{document}